\documentclass[letterpaper, 10 pt, conference]{ieeeconf}
\usepackage{amsmath}
\usepackage{amssymb}
\usepackage{url}
\usepackage{graphicx}
\usepackage{booktabs}
\usepackage{multirow}
\usepackage{tabularx}
\usepackage{balance}
\usepackage{makecell}

\newcommand{\rev}[1]{#1}
\newcommand{\revz}[1]{#1}

\IEEEoverridecommandlockouts
\title{\LARGE \bf
One Image is All You Need: Agentic One-Shot Image Generation via Text-Based World Models for Long-Tail Spatial Perception
}

\author{Keqin Zeng,
        Shuting Su,
        Shihao Lin,
        Rui~Zhao,
        Ziyue~Li$^*$
\thanks{Keqin Zeng is with Tsinghua University. Shuting Su is with SenseTime Research. Shihao Lin is with Sun Yat-Sen University. Rui~Zhao is with SenseTime Research. E-mail: zhaorui@sensetime.com.}%
\thanks{Ziyue Li is with Technical University of Munich, Heilbronn Data Science Center, and Munich Data Institute. E-mail: ziyue.li@tum.de.}%
\thanks{$^*$Corresponding author: Ziyue Li.}
}

\begin{document}

\maketitle
\thispagestyle{empty}
\pagestyle{empty}

\begin{abstract}
Reliable spatial decision automation, such as autonomous driving and maritime surveillance, critically depends on robust visual perception. However, real-world spatiotemporal data exhibits severe heterogeneity, often manifesting as extreme long-tail distributions for safety-critical scenarios. This data scarcity induces dataset shift that degrades detection performance and pose safety risks. While synthetic data generation offers a potential solution, existing generative approaches, such as diffusion models and Generative Adversarial Networks (GANs), often lack explicit spatial grounding and structural constraints, resulting in spatial and physical inconsistencies in generated scenes. To address these challenges, we introduce WMGen-v1, an agentic text-based world model framework for long-tail spatial data generation. WMGen-v1 employs a Large Vision-Language Model (LVLM) to construct a structured scene representation from a single reference image, while a Large Language Model (LLM) performs guidance-based scene expansion under physical plausibility and commonsense constraints. Subsequently, conditioned on the structured semantic representations produced by this reasoning process, a diffusion model generates diverse and physically grounded long-tail training data. Experiments on internal industrial datasets, ROADWork, and LaRS benchmarks demonstrate that WMGen-v1 outperforms baseline approaches. \revz{Notably, detectors trained solely on WMGen-v1 synthetic data approach real-only performance on aggregate dataset-level metrics, highlighting its potential to alleviate long-tail data scarcity for downstream spatial perception.}
\end{abstract}

\section{INTRODUCTION}
Modern automation systems are increasingly deployed in complex and open-world environments. In spatial automation domains, such as autonomous driving and unmanned surface vehicles, the reliability of downstream decision-making is strongly influenced by the robustness of upstream visual perception. However, visual data collected in these environments is typically highly heterogeneous, naturally resulting in a long-tail distribution in which safety-relevant rare scenarios and object categories are severely underrepresented. This imbalance weakens the perception capability of automation systems on rare but critical cases, increasing the risk of missed detections in rare but safety-critical cases.

Collecting real-world data for such rare scenarios is costly, time-consuming, and often impractical. Synthetic data generation therefore provides a promising alternative for improving long-tail perception. Nevertheless, existing generation methods still face important limitations. Single-image GANs approaches, such as One-Shot GAN \cite{sushko2021sho}, often struggle to preserve semantic fidelity and structural consistency under complex scene variations. Meanwhile, diffusion-based generation pipelines without explicit spatial grounding are prone to semantic drift, layout inconsistencies, and physically implausible details, which limits their usefulness for downstream perception tasks.

To address these challenges, we propose WMGen-v1, an agentic data generation framework that serves as a text-based world model for long-tail visual perception. WMGen-v1 leverages the complementary strengths of LVLMs, LLMs, and diffusion models. Given a single reference image, an LVLM first extracts geometric, semantic, and spatial constraints from the observed scene. An LLM then performs structured scene reasoning to enrich the scene with plausible long-tail variations under physical and contextual priors. Finally, a diffusion model synthesizes reference-conditioned images from the resulting semantic descriptions, producing physically grounded and spatially coherent long-tail samples for downstream detector training. In this way, WMGen-v1 offers a practical generation pipeline that requires only one user instruction and one reference image, while improving the controllability and realism of synthetic data for safety-critical spatial scenarios.

\begin{figure}[t]
    \centering
    \includegraphics[width=\linewidth]{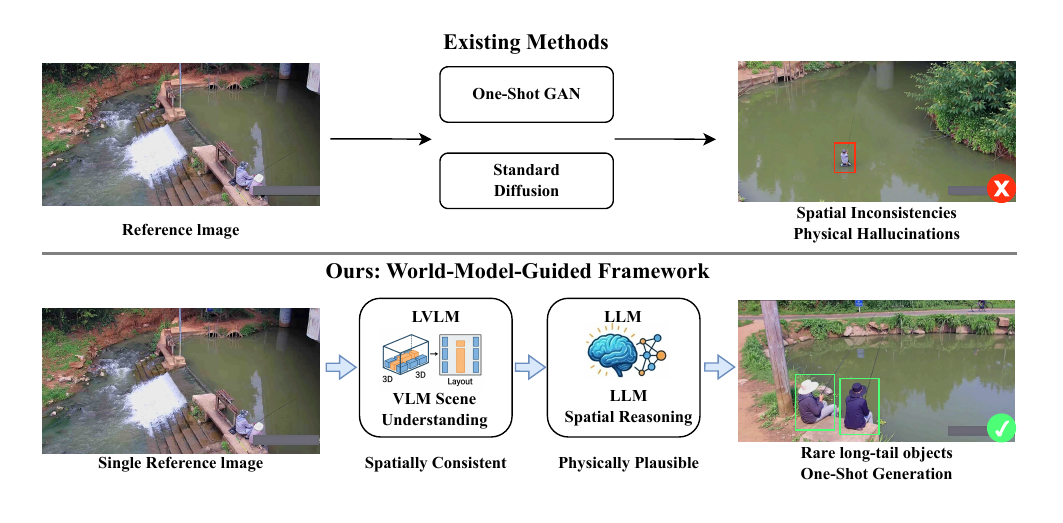}
    \vspace{-20pt}
    \caption{Conceptual comparison of long-tail data generation. Unlike conventional generative methods (top) that suffer from spatial and physical inconsistencies when synthesizing complex scenes, WMGen-v1 (bottom) \rev{better preserves spatial coherence and physical plausibility in our evaluated settings}, yielding high-fidelity data for long-tail perception tasks.}
    \vspace{-10pt}
    \label{fig:abstract}
\end{figure}

The main contributions of this paper are:
\begin{itemize}
    \item We propose WMGen-v1, an agentic text-based world-model framework for mitigating long-tail data scarcity in spatial perception. The framework unifies structured scene modeling, world-model expansion, and reference-conditioned image synthesis.
    
    \item We introduce a multimodal scene modeling pipeline that constructs structured scene constraints from a single reference image and performs guidance-based world-model expansion under physical plausibility and commonsense constraints, while preserving key structural anchors during generation.
    
    \item Evaluations on ROADWork \cite{ghosh2023roa}, LaRS \cite{ust2023lar}, and internal datasets demonstrate that WMGen-v1 improves downstream detection performance over baseline synthetic data generation methods. Notably, detectors trained solely on synthetic data generated by WMGen-v1 remain competitive with real-data training.
\end{itemize}

\section{RELATED WORKS}
\subsection{Long-tail Object Detection}
Robust object detection is the cornerstone of spatial automation, yet automation systems frequently face unforeseen operational disruptions and highly variable environments. \revz{Existing long-tail detection methods can be grouped into data-level augmentation, which increases sample diversity by transforming or synthesizing training samples \cite{guo2025srm,yang2023lon}; loss reweighting, which adjusts the contribution of imbalanced classes during optimization \cite{tan2020equ,hyuncho2022lon}; and decoupled representation learning, which mitigates classifier bias by separating representation learning from classifier rebalancing \cite{du2024pro,cong2024dec}.} However, these methods remain constrained by available training data and cannot generate samples for previously unseen scene configurations. In practical deployments, multi-agent or decentralized control architectures improve adaptability, but agents typically operate within predefined capability ranges set at initialization. When encountering rare, out-of-distribution spatiotemporal contexts, narrowly trained perception models fail to identify critical obstacles, forcing the system to act on flawed environmental models and leading to potential decision failures. Addressing the long-tail perception deficit through synthetic generation is therefore critical to expanding the safe operational boundaries of automated systems.

\subsection{Synthetic Data Generation}
Synthetic data generation mitigates the scarcity of rare events. Traditional GAN-based approaches, exemplified by One-Shot GAN \cite{sushko2021sho}, learn distributions from limited samples but often suffer from training instability, pixel-level overfitting, and limited capacity to generalize complex spatiotemporal relationships. Recent diffusion models, such as SDXL \cite{podell2023sdx} and Wuerstchen \cite{pernias2023wue}, deliver superior image synthesis quality and diversity. Despite these advances, text-to-image generation lacks grounded physical constraints, leading to spatial inconsistencies and physical implausibility in complex multi-object scenes. These hallucinations, such as floating or overlapping objects, undermine downstream reliability. \revz{To address this, controllable diffusion models have incorporated geometric constraints and structural guidance for data augmentation in downstream perception tasks \cite{chen2024geo,fang2024dat}. For instance, GeoDiffusion encodes geometric conditions (e.g., 3D layout) into text prompts for generalized layout-to-image generation \cite{chen2024geo}, while ControlAug employs visual priors and category-calibrated filtering to augment object detection data \cite{fang2024dat}. Although these methods enhance generator-side controllability, they heavily rely on explicit, predefined control inputs and typically require additional training or adaptation of the diffusion components.} 

\subsection{LLM/VLM Scene Reasoning}
The emergence of LLMs and LVLMs has revolutionized cognitive reasoning and task automation. \revz{Modern VLMs improve multimodal perception \cite{li2026min,fu2024bli}, while spatially oriented VLM studies examine whether such models can encode geometric or viewpoint information \cite{xu2024lla}. In parallel, LLM-based planning methods translate natural-language instructions into structured plans or layout guidance for downstream generation \cite{xia2023llm,chen2026spa}.} However, existing methods typically use these models solely for prompt generation or multimodal planning, without explicitly modeling scene structure for synthetic dataset generation. To address this limitation, our approach leverages LVLMs to convert complex visual structures into precise textual descriptions, enabling explicit spatiotemporal reasoning while strengthening visual grounding to reduce hallucinations. By fusing representations within a shared discrete semantic space, these foundation models can guide generative processes to produce long-tail data that remains logically rigorous, contextually accurate, and bound by physical priors.

\section{ALGORITHM FRAMEWORK}

\begin{figure*}[htbp]
    \centering
    \includegraphics[width=\textwidth]{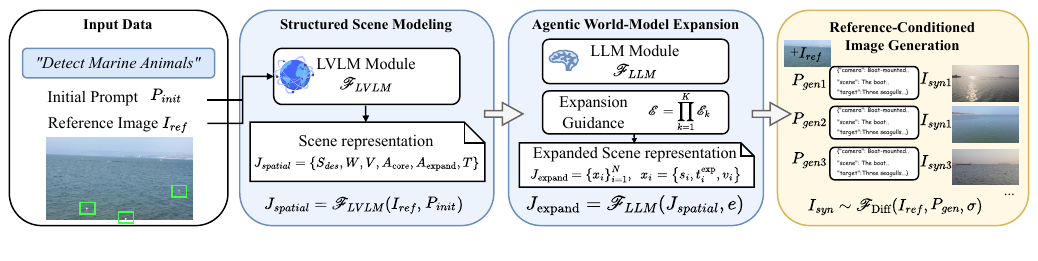}
    \vspace{-30pt}
    \caption{Overview of the proposed WMGen-v1 framework. The pipeline consists of three stages: (1) Structured Scene Modeling, where an LVLM constructs a structured scene representation from a single reference image; (2) Agentic World-Model Expansion, where an LLM performs guidance-based scene expansion under physical plausibility and commonsense constraints; and (3) Reference-Conditioned Image Generation, where a diffusion model synthesizes structurally consistent long-tail samples conditioned on both the reference image and the expanded prompts.}
    \label{fig:framework}
\end{figure*}

We present WMGen-v1, an agentic text-based \textbf{W}orld \textbf{M}odel framework for long-tail spatial data \textbf{Gen}eration. As shown in Figure~\ref{fig:framework}, the system decomposes the long-tail generation task into three phases: Structured Scene Modeling, Agentic World-Model Expansion, and Reference-Conditioned Image Synthesis.
\rev{The supplementary material includes a case study that follows a reference image through the complete generation pipeline, covering $I_{ref}$, $J_{spatial}$, $J_{expand}$, $P_{gen}$, and the resulting synthesized image $I_{syn}$.}

\subsection{Problem Formulation}
In real-world spatial automation systems, let the target physical environment be represented by a dataset $\mathcal{D}$. Due to natural long-tail distributions, $\mathcal{D}$ can be partitioned into head classes $\mathcal{C}_{head}$ and rare tail classes $\mathcal{C}_{tail}$.

Given a limited reference set $\mathcal{I}_{ref} \subset \mathbb{R}^{H \times W \times 3}$, where each image $I_{ref} \in \mathcal{I}_{ref}$ contains long-tail objects $c \in \mathcal{C}_{tail}$, our objective is to synthesize a structurally consistent and physically grounded augmented dataset $\mathcal{D}_{syn}$ to mitigate dataset shift. We frame this as a stochastic conditional mapping $\Phi$. Given a guidance condition $e \in \mathcal{E}$ from a predefined guidance space, the generation of a novel long-tail image $I_{syn}$ is formulated as:
\begin{equation}
I_{syn} \sim \Phi(\cdot \mid I_{ref}, e).
\end{equation}

\subsection{Structured Scene Modeling}
To reduce spatial inconsistency and physical implausibility in the generated images, we first employ an LVLM to construct a structured scene model from the reference image.

Let $P_{init}$ denote the initial prompt that instructs the LVLM to analyze scene composition, target semantics, camera properties, and spatial structure. Given $I_{ref}$ and $P_{init}$, the LVLM produces a structured scene representation under a predefined schema, denoted as $J_{spatial}$:
\begin{equation}
J_{spatial} = \mathcal{F}_{LVLM}(I_{ref}, P_{init}).
\end{equation}
We represent $J_{spatial}$ as:
\begin{equation}
J_{spatial} = \{S_{des}, W, V, A_{core}, A_{expand}, T\},
\end{equation}
where $S_{des}$ denotes grounded scene descriptions, $W$ is the global scene abstraction, $V$ represents camera and viewpoint constraints, $A_{core}$ and $A_{expand}$ denote the core and expandable scene regions, respectively, and $T=\{t_i\}_{i=1}^{M}$ is a set of target-centric descriptors. Each $t_i = \{n_i, l_i, d_i, m_i, u_i\}$ encapsulates fine-grained target information, including its name, task label, semantic description, grounding index, and visual attributes.

\subsection{Agentic World-Model Expansion}
To expand the generated data toward diverse out-of-distribution scenarios, we utilize an LLM to perform agentic expansion over the structured scene model.

We define an expansion guidance space $\mathcal{E}$, which specifies a set of semantic dimensions for scene expansion, such as weather, illumination, scene context, and object-level attributes. During the expansion phase, the LLM expands the scene under guidance $e \in \mathcal{E}$. Conditioned on the structured scene anchor $J_{spatial}$ and the guidance condition $e$, the LLM outputs a schema-constrained expanded scene representation, denoted as $J_{expand}$:
\begin{equation}
J_{expand} = \mathcal{F}_{LLM}(J_{spatial}, e).
\end{equation}
We represent the expanded scene state as a set of scene descriptors:
\begin{equation}
J_{expand} = \{x_i\}_{i=1}^{N}, \quad x_i = \{s_i, t_i^{exp}, v_i\},
\end{equation}
where $s_i$ denotes the expanded scene description, $t_i^{exp}$ specifies the target object description under the new environment, and $v_i$ provides the corresponding camera-consistent rendering instruction.

The final generation prompt is then constructed from the expanded scene representation:
\begin{equation}
P_{gen} = \mathcal{G}(J_{expand}).
\end{equation}
Unlike pure prompt rewriting, $\mathcal{F}_{LLM}$ performs constrained scene expansion under explicit physical plausibility and commonsense guidance, ensuring that the target object remains compatible with the sampled environment while preserving the key structural cues inherited from $J_{spatial}$.

\subsection{Reference-Conditioned Image Generation}
Finally, the constructed prompt $P_{gen}$ and the original reference image $I_{ref}$ are fed into a diffusion model for reference-conditioned synthesis:
\begin{equation}
I_{syn} \sim \mathcal{F}_{Diff}(I_{ref}, P_{gen}, \sigma),
\end{equation}
where $\sigma$ denotes the diffusion generation control parameter. This reference-conditioned mechanism guides the diffusion model to preserve the structural anchors and visual configuration of $I_{ref}$ while rendering new contextual variations specified by $P_{gen}$. As a result, the synthesized image $I_{syn}$ maintains spatial consistency for downstream perception tasks while introducing substantial environmental diversity.

\section{EXPERIMENTS}
\subsection{Dataset}
We evaluate WMGen-v1 on three datasets that exhibit pronounced long-tail characteristics across safety-critical spatial perception scenarios, including autonomous driving in construction zones, maritime surveillance, and industrial monitoring.

\subsubsection{ROADWork Dataset} 
The ROADWork dataset \cite{ghosh2023roa} focuses on perception and navigation in construction zones for autonomous driving. It contains 5,318 training images and 2,098 validation images annotated with various traffic-control objects and construction-related entities. Similar to many real-world safety scenarios, ROADWork exhibits a highly skewed class distribution. Common objects such as traffic \textit{Cone}s appear in 2,993 images, while critical yet rare categories including \textit{Police Officer} and \textit{Bike Lane} occur in fewer than 90 images. Such long-tail characteristics make the dataset a challenging benchmark for safety-critical object detection in complex road environments.

\subsubsection{LaRS Dataset} 
The LaRS dataset \cite{ust2023lar} is a maritime obstacle detection benchmark covering diverse environments including lakes, rivers, and coastal waters. It contains 2,605 training images and 198 validation images, with annotations spanning both obstacle and environmental classes. The dataset exhibits a pronounced long-tail distribution. While head classes such as \textit{Water} and \textit{Sky} dominate the dataset, several safety-critical obstacle categories are severely underrepresented. For instance, rare classes such as \textit{Float} and \textit{Animal} appear in only 20 and 72 training images, respectively. This imbalance makes LaRS particularly suitable for evaluating long-tail perception under maritime surveillance conditions.

\subsubsection{Internal Industrial Monitoring Dataset}
In addition to public benchmarks, we also evaluate WMGen-v1 on an internal industrial monitoring dataset collected from real-world surveillance systems deployed in safety-critical environments. This private dataset contains monitoring images captured in practical operational scenarios, including reservoir surveillance and industrial safety monitoring tasks such as fishing detection and smoke detection. Due to confidentiality constraints, this dataset is partially used for qualitative evaluation to assess the realism and practical utility of the generated images.

To alleviate the severe long-tail imbalance, we selectively generate a limited number of synthetic data to augment the most underrepresented tail classes. This design allows us to evaluate whether a small amount of targeted synthetic data can effectively mitigate long-tail imbalance. For the LaRS dataset, we generated additional images for tail class including \textit{Animal} and \textit{Float}. For the Roadwork dataset, we targeted \textit{Police Officer, Police Vehicle, Bike Lane}, and \textit{Other Roadwork Objects}. Detailed statistics for the original and generated dataset can be found in Table \ref{tab:dataset_stats}.

{\setlength{\textfloatsep}{4pt plus 1pt minus 1pt}
\begin{table}[t]
\caption{Statistics of the LaRS and Roadwork Datasets}
\label{tab:dataset_stats}
\centering
\footnotesize
\setlength{\tabcolsep}{3pt}
\renewcommand{\arraystretch}{0.92}
\setlength{\aboverulesep}{0.5pt}
\setlength{\belowrulesep}{0.5pt}
\begin{tabular}{l l r r c}
\toprule
\textbf{Dataset} & \textbf{Class Name} & \textbf{Train (Real)} & \textbf{Val} & \textbf{Generated} \\
\midrule
\multirow{18}{*}{\textbf{Roadwork}} 
& Police Officer & 69 & 44 & 100 \\
& Police Vehicle & 87 & 45 & 100 \\
& Cone & 2993 & 1301 & - \\
& Fence & 1319 & 646 & - \\
& Drum & 949 & 440 & - \\
& Barricade & 1616 & 579 & - \\
& Barrier & 1365 & 480 & - \\
& Work Vehicle & 2790 & 1283 & - \\
& Vertical Panel & 1046 & 240 & - \\
& Tubular Marker & 1354 & 788 & - \\
& Arrow Board & 587 & 158 & - \\
& Bike Lane & 61 & 26 & 40 \\
& Work Equipment & 464 & 221 & - \\
& Worker & 1146 & 545 & - \\
& Other Roadwork Objects & 48 & 14 & 40 \\
& Temp Traffic Msg Board & 162 & 69 & - \\
& Temp Traffic Sign & 2384 & 758 & - \\
\cmidrule{2-5}
& \textbf{Total Images} & \textbf{5318} & \textbf{2098} & \textbf{280} \\
\midrule
\multirow{12}{*}{\textbf{LaRS}} 
& Static Obstacle & 2533 & 186 & - \\
& Water & 2605 & 198 & - \\
& Sky & 2492 & 176 & - \\
& Boat/ship & 1465 & 132 & - \\
& Row boats & 228 & 21 & - \\
& Paddle board & 110 & 16 & - \\
& Buoy & 730 & 48 & - \\
& Swimmer & 120 & 13 & - \\
& Animal & 72 & 4 & 40 \\
& Float & 20 & 3 & 20 \\
& Other & 279 & 37 & - \\
\cmidrule{2-5}
& \textbf{Total Images} & \textbf{2605} & \textbf{198} & \textbf{60} \\
\bottomrule
\end{tabular}
\end{table}}

\subsection{Detection Models and Evaluation Metrics}

\textit {1) Detection Models:}
To evaluate the effectiveness of the generated long-tail data, we conduct experiments using two widely adopted object detection architectures representing different detection paradigms: a one-stage detector and a two-stage detector.

For the one-stage setting, we use YOLO11s \cite{hassani2026yolo}, a lightweight YOLO-based detector designed for efficient dense prediction and real-time deployment. For the two-stage setting, we adopt Faster R-CNN with a ResNet-152 backbone \cite{ren2015faster}, which employs a Region Proposal Network (RPN) to generate candidate regions followed by region-wise classification and bounding-box regression.

Evaluating WMGen-v1 on both architectures allows us to verify that the improvements brought by the generated data are not tied to a specific detection framework and generalize across different detection paradigms.

\textit {2) Evaluation Metrics:}
Following the MS COCO evaluation protocol \cite{chen2015microsoft}, we report mean Average Precision (mAP) as the primary evaluation metric. Specifically, we report mAP@50 and the stricter mAP@50:95 averaged across IoU thresholds from 0.5 to 0.95.

Given the severe class imbalance in both datasets, we further report per-class mAP@50 to explicitly measure performance improvements on rare tail categories. 

\subsection{Baselines and Implementation Details}

\textit {1) Baseline Configurations:}
To evaluate the effectiveness of WMGen-v1, we establish five experimental settings: \textit{Real-only}, \textit{Real + Baseline 1}, \textit{Real + Baseline 2}, \textit{Ours-only}, and \textit{Real + WMGen-v1}. The baselines are strategically selected to represent existing generative paradigms, ensuring fair comparisons under identical reference inputs and generated sample quantities. \textit{Ours-only} evaluates whether models trained purely on synthetic data generated by WMGen-v1 can achieve competitive performance compared with models trained on real data.

\begin{itemize}
    \item \textbf{Baseline 1 (LLM-Augmented Diffusion).} 
    This baseline represents the prevalent generation paradigm \cite{labs2025flux}. It employs the same reference-conditioned diffusion model as WMGen-v1 but replaces our structured, dual-agent scene reasoning with a standard LLM-based prompt rewriting strategy. The LLM simply generates unconstrained textual variations from the user input to guide the diffusion model. Comparing against this setup highlights the critical necessity of our explicit multimodal spatial deduction and strict physical prior injection.
    
    \item \textbf{Baseline 2 (One-Shot GAN).} 
    This baseline adopts a representative one-shot image generation approach that learns internal patch distributions also from a single reference image \cite{sushko2021sho}. Unlike WMGen-v1, which leverages multimodal reasoning and diffusion-based generation, this method relies solely on internal image statistics to synthesize new samples. 
\end{itemize}

\textit {2) Implementation Details:}
All experiments were conducted on NVIDIA V100 GPUs. All methods share the same reference images and generate the same number of augmented samples for the target tail classes to ensure a fair comparison.

The backbone of the LVLM and LLM are ChatGPT-4o and ChatGPT-5.2. For data generation, WMGen-v1 employs FLUX.1 Kontext Pro as the underlying diffusion model, which we selected after preliminary experiments for its favorable balance between generation quality and computational cost. \revz{For reproducibility, the full schemas, prompt templates, retry protocol, and implementation details are included in the supplementary material.}

To annotate the generated images, we adopt Grounding DINO \cite{liu2024grounding}, an open-vocabulary object detector, for automatic label generation. Specifically, all synthetic samples are pseudo-labeled using Grounding DINO with a confidence threshold of 0.3. This unified auto-annotation strategy is applied to all generated data across both datasets, ensuring consistency in the labeling process.

For Baseline 2, we follow the original training configuration. Empirical observations show that training beyond 60k iterations yields negligible improvements compared with longer runs such as 150k we tried before. Therefore, we adopt 60k iterations for generation on both datasets to balance generation quality and computational efficiency.

For downstream evaluation, YOLO11s and Faster R-CNN are trained for 150 epochs using their default training configurations. No task-specific hyperparameter tuning is applied, ensuring that performance differences primarily reflect the quality and effectiveness of the generated training data.

\subsection{Ablation Study}

To validate the necessity of the proposed architecture, we conduct an ablation study to isolate the contribution of the LLM reasoning module. Specifically, we introduce the \textit{w/o LLM} (LVLM-only) configuration. In this variant, the diffusion model is conditioned solely on the raw spatial layout ($J_{spatial}$) extracted by the LVLM, without the physical priors and logical scene modeling provided by the LLM.

All other components and training settings remain identical to the full pipeline, ensuring that the observed performance differences are attributable solely to the removal of the LLM reasoning module.

\subsection{Qualitative Results}
\begin{figure*}[htbp]
    \centering
    \setlength{\tabcolsep}{2pt} 
    \begin{tabular}{@{} c ccccc @{}}
        
        & \small Reference Image & \small Baseline 1 & \small Baseline 2 & \small WMGen-v1 w/o LLM & \small \textbf{WMGen-v1} \\
        \addlinespace[4pt] 
        
        \raisebox{0.1cm}{\rotatebox{90}{\scriptsize {Internal Dataset}}} &
        \includegraphics[width=0.188\textwidth]{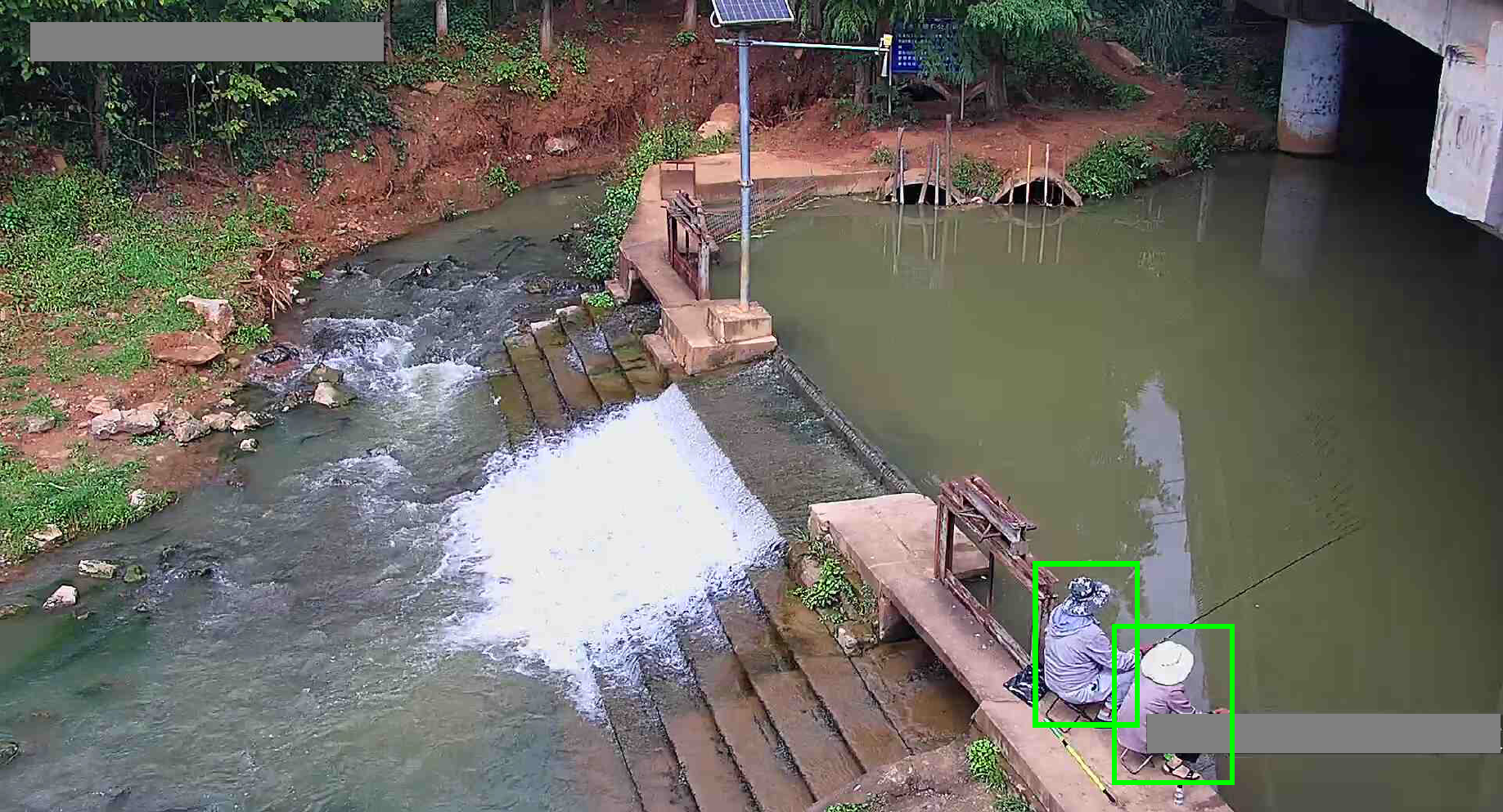} &
        \includegraphics[width=0.188\textwidth]{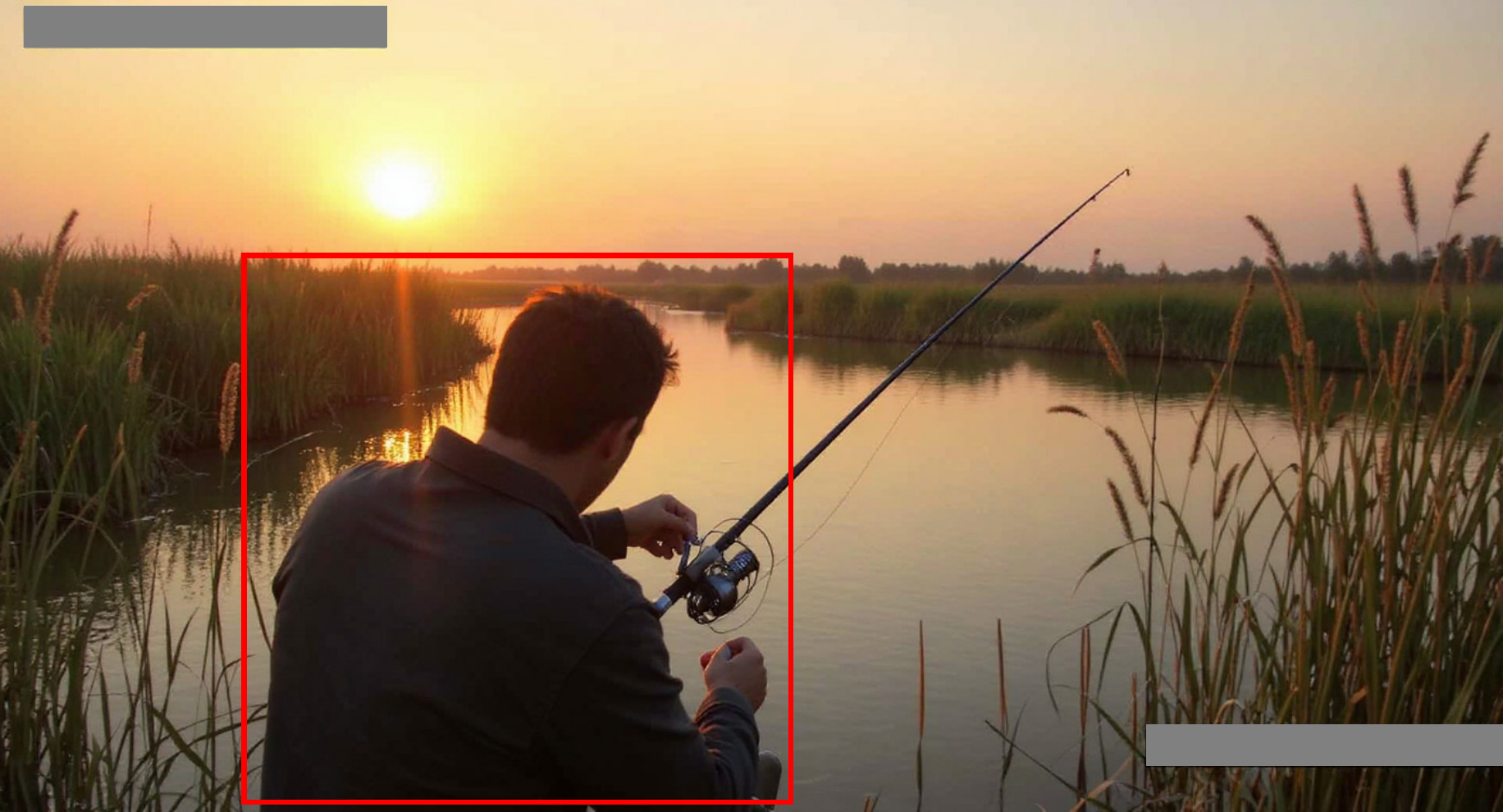} &
        \includegraphics[width=0.188\textwidth]{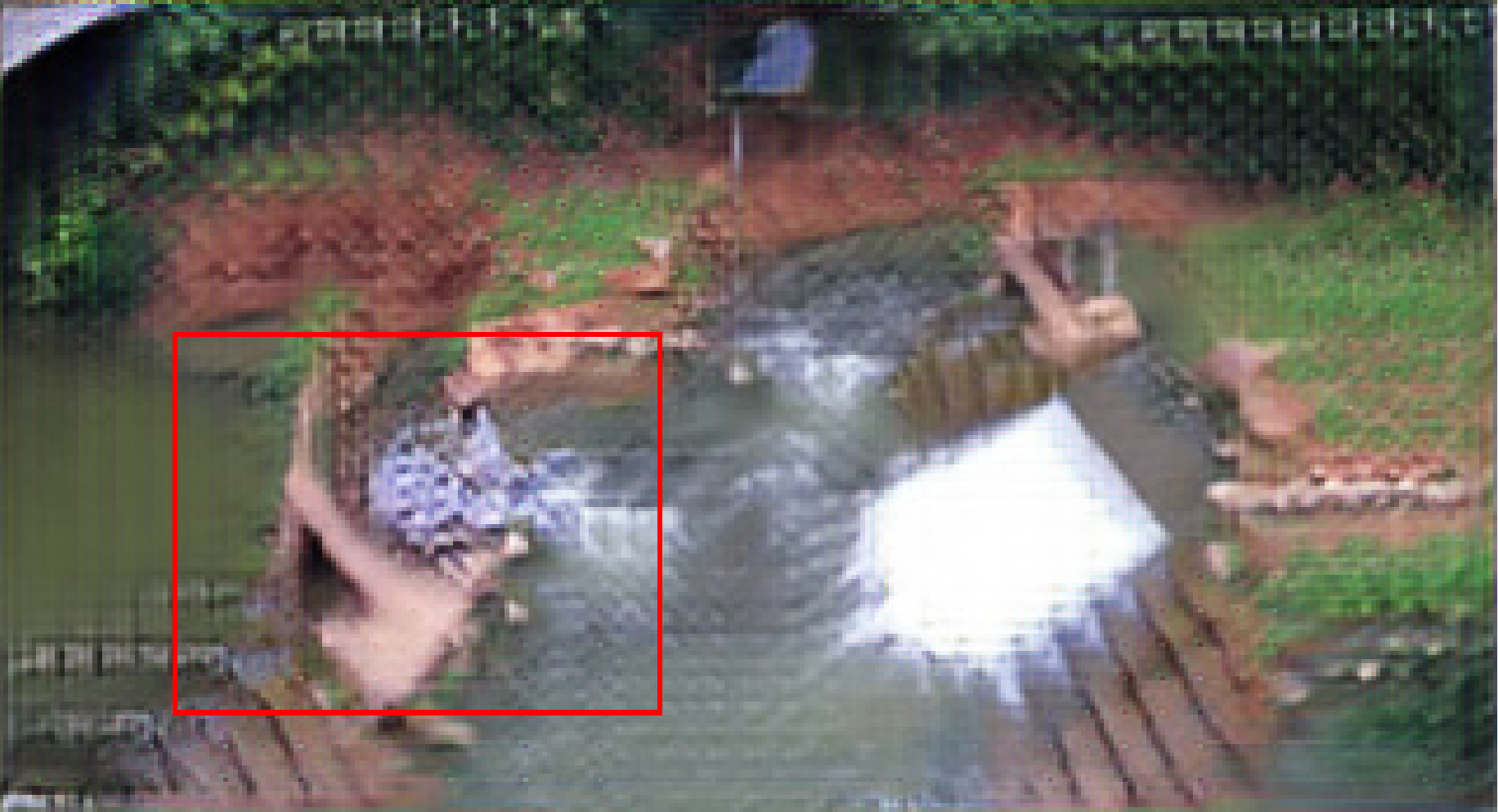} &
        \includegraphics[width=0.188\textwidth]{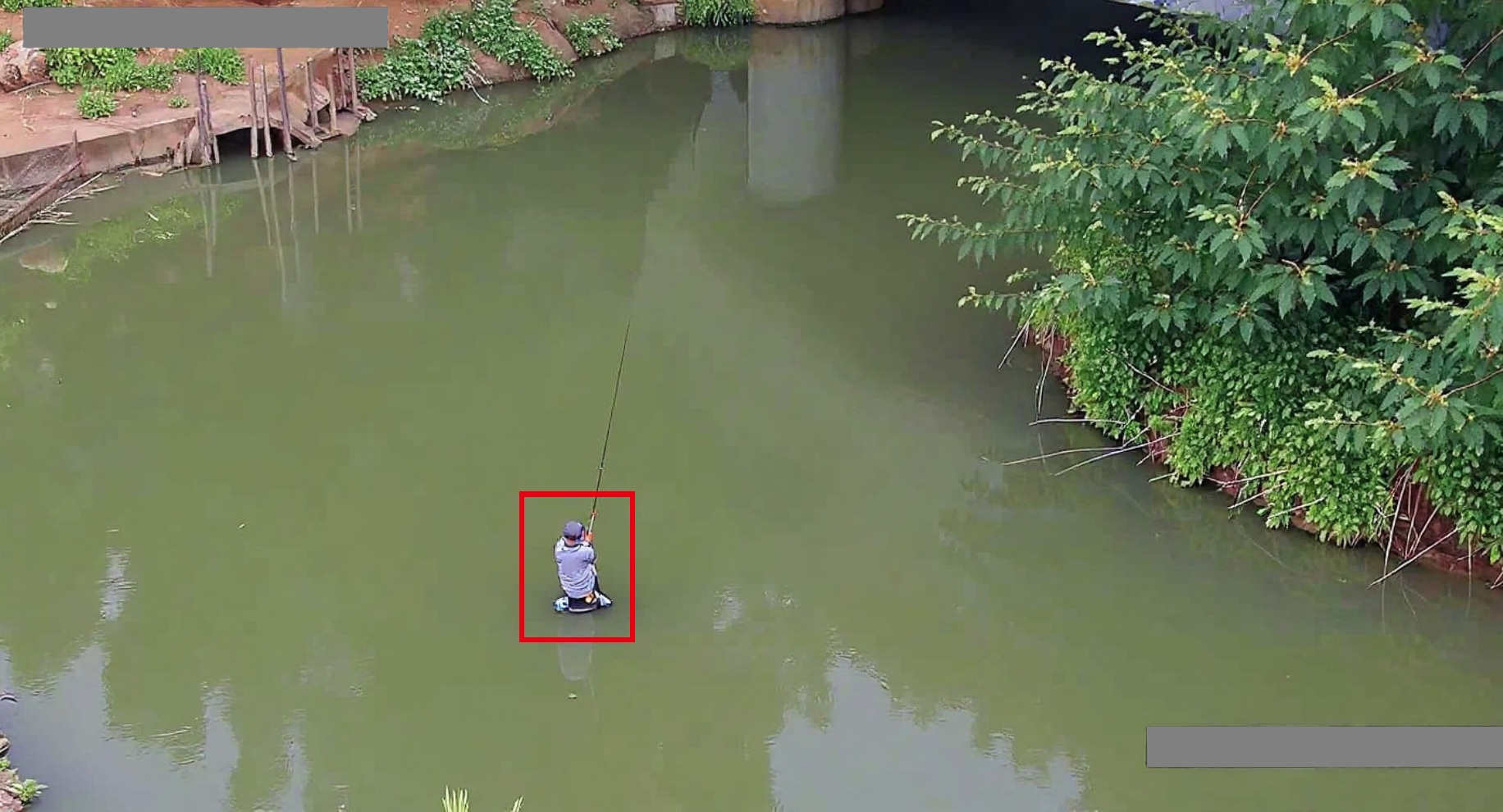} &
        \includegraphics[width=0.188\textwidth]{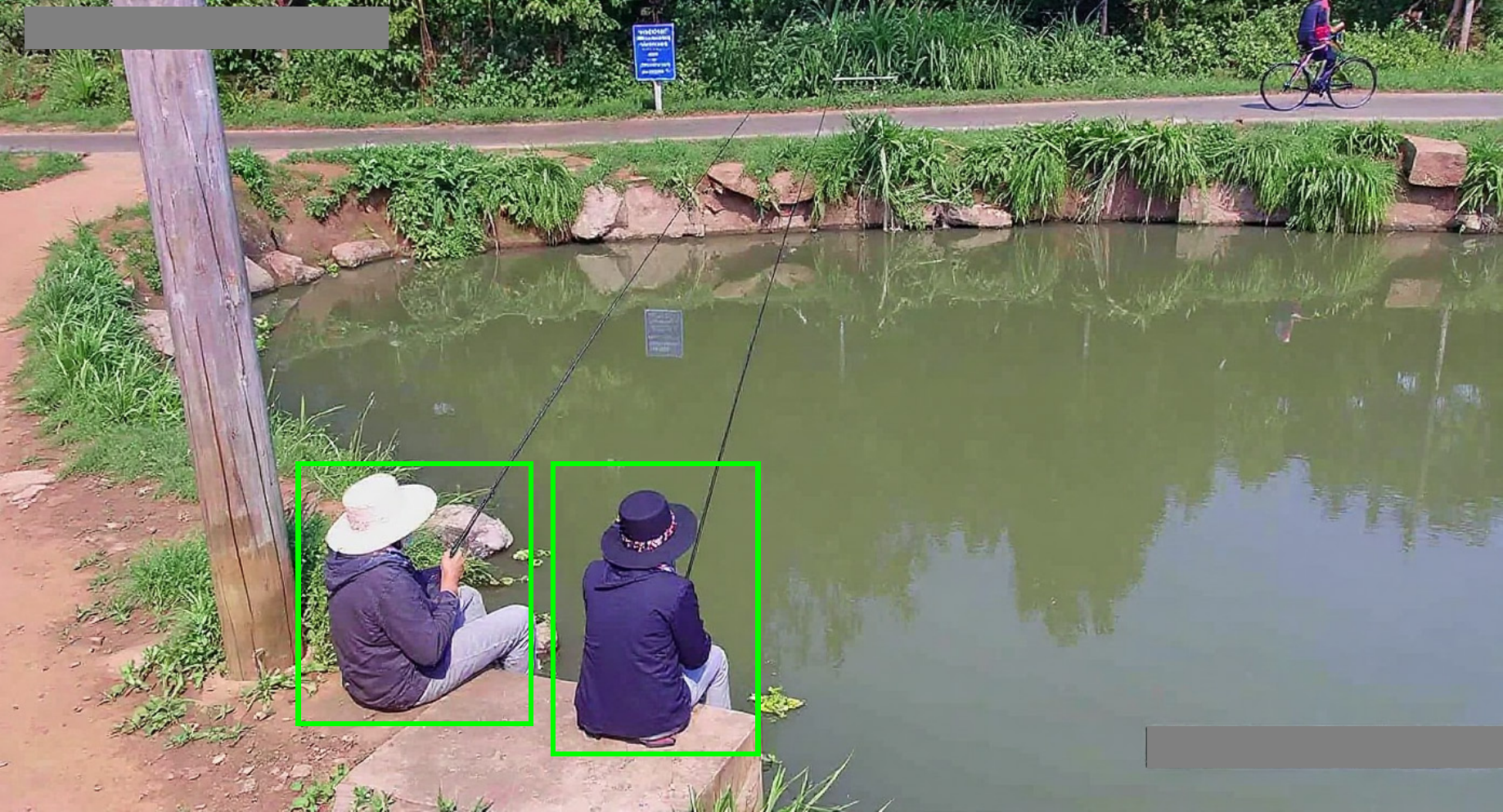} \\
        \addlinespace[2pt]
        
        \raisebox{0.3cm}{\rotatebox{90}{\scriptsize {ROADWork}}} &
        \includegraphics[width=0.188\textwidth]{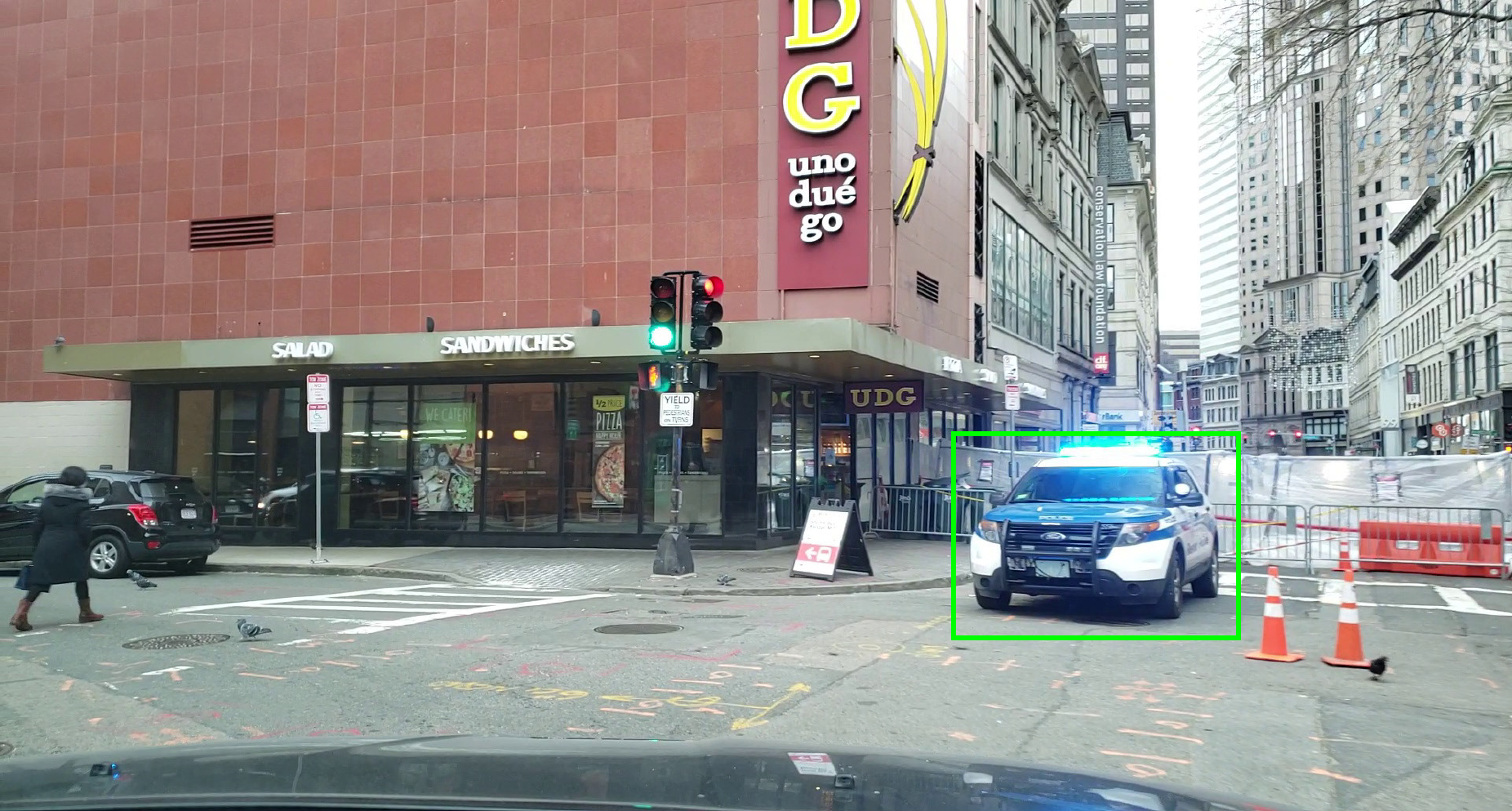} &
        \includegraphics[width=0.188\textwidth]{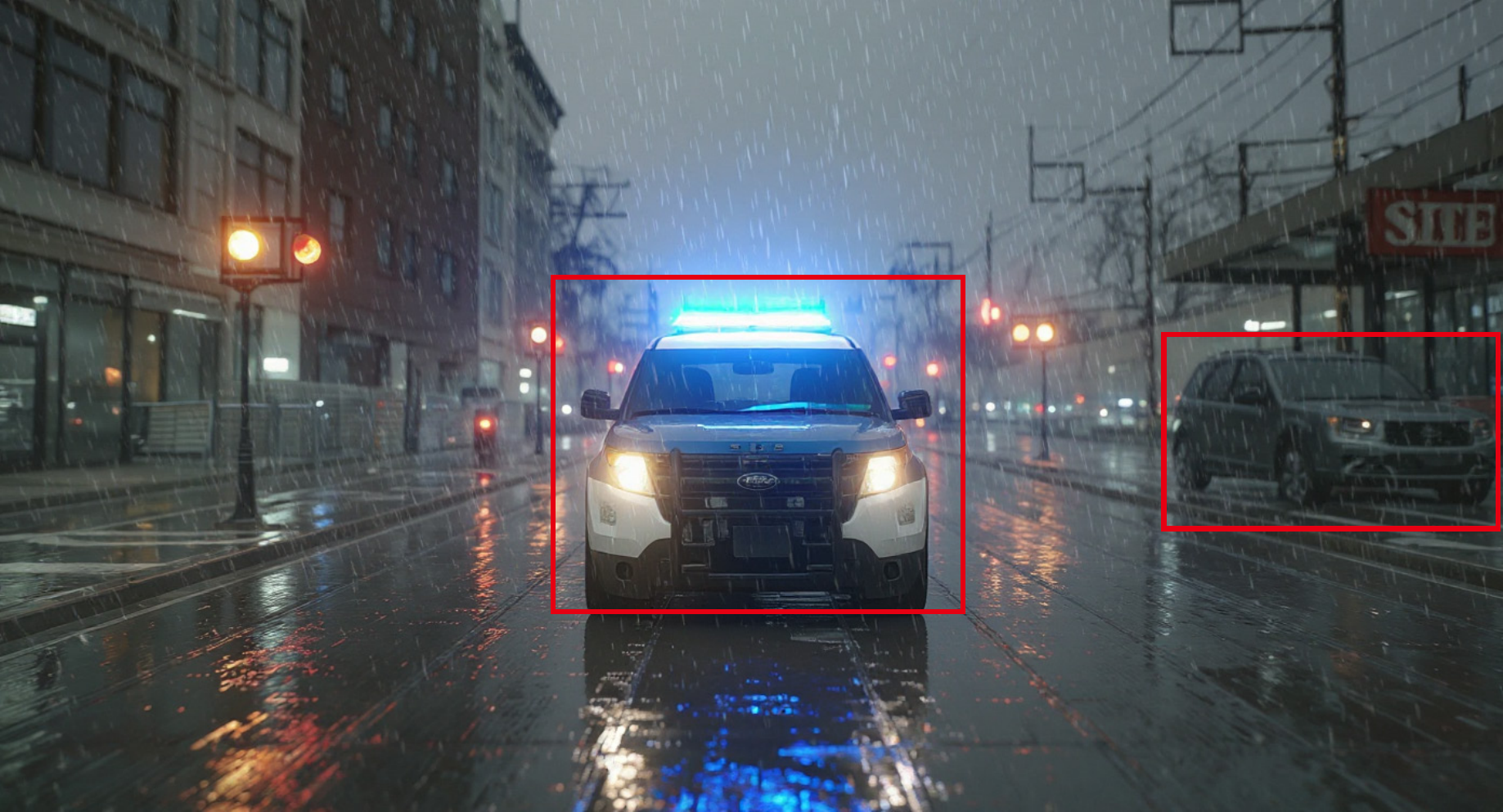} &
        \includegraphics[width=0.188\textwidth]{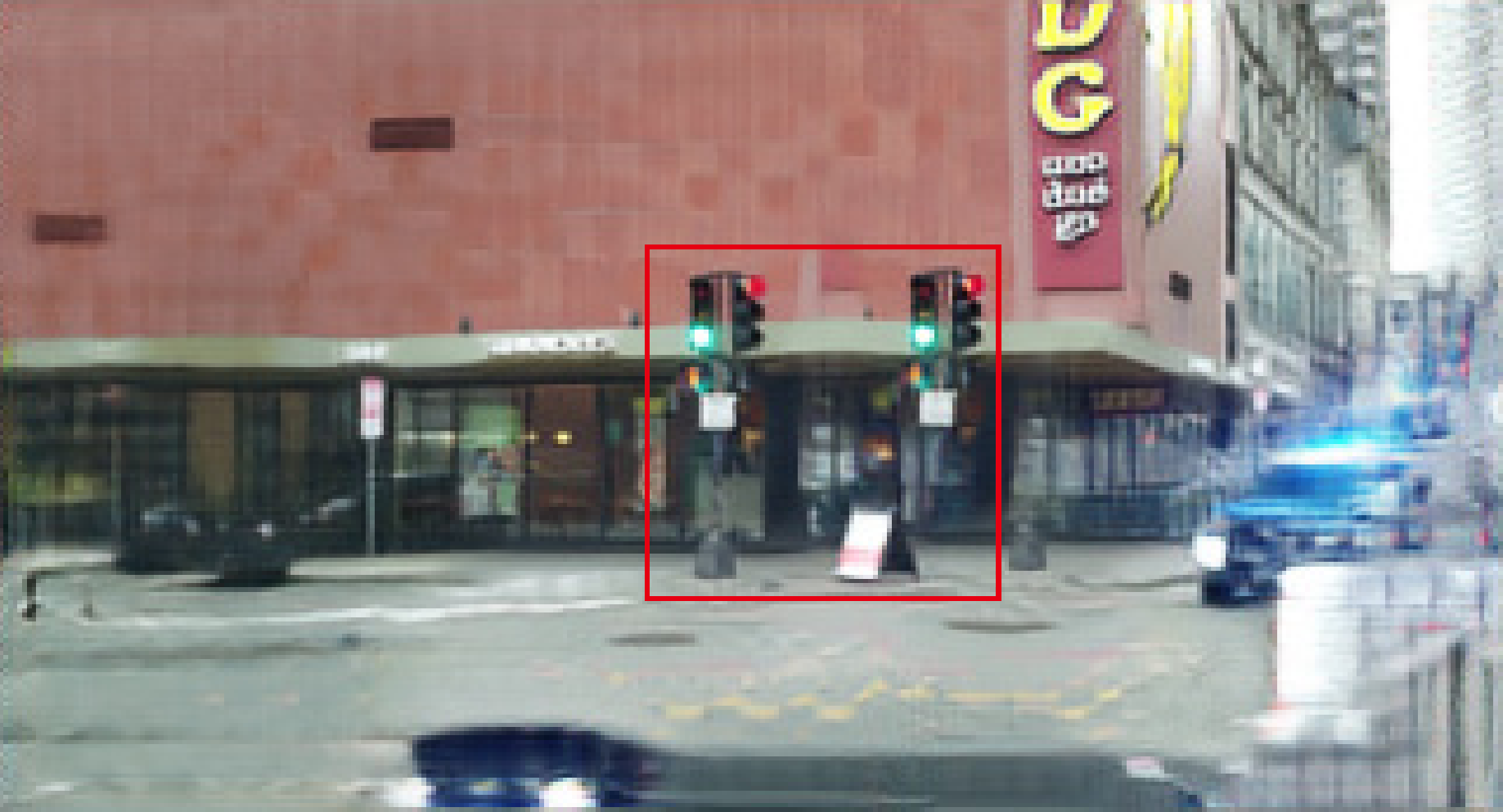} &
        \includegraphics[width=0.188\textwidth]{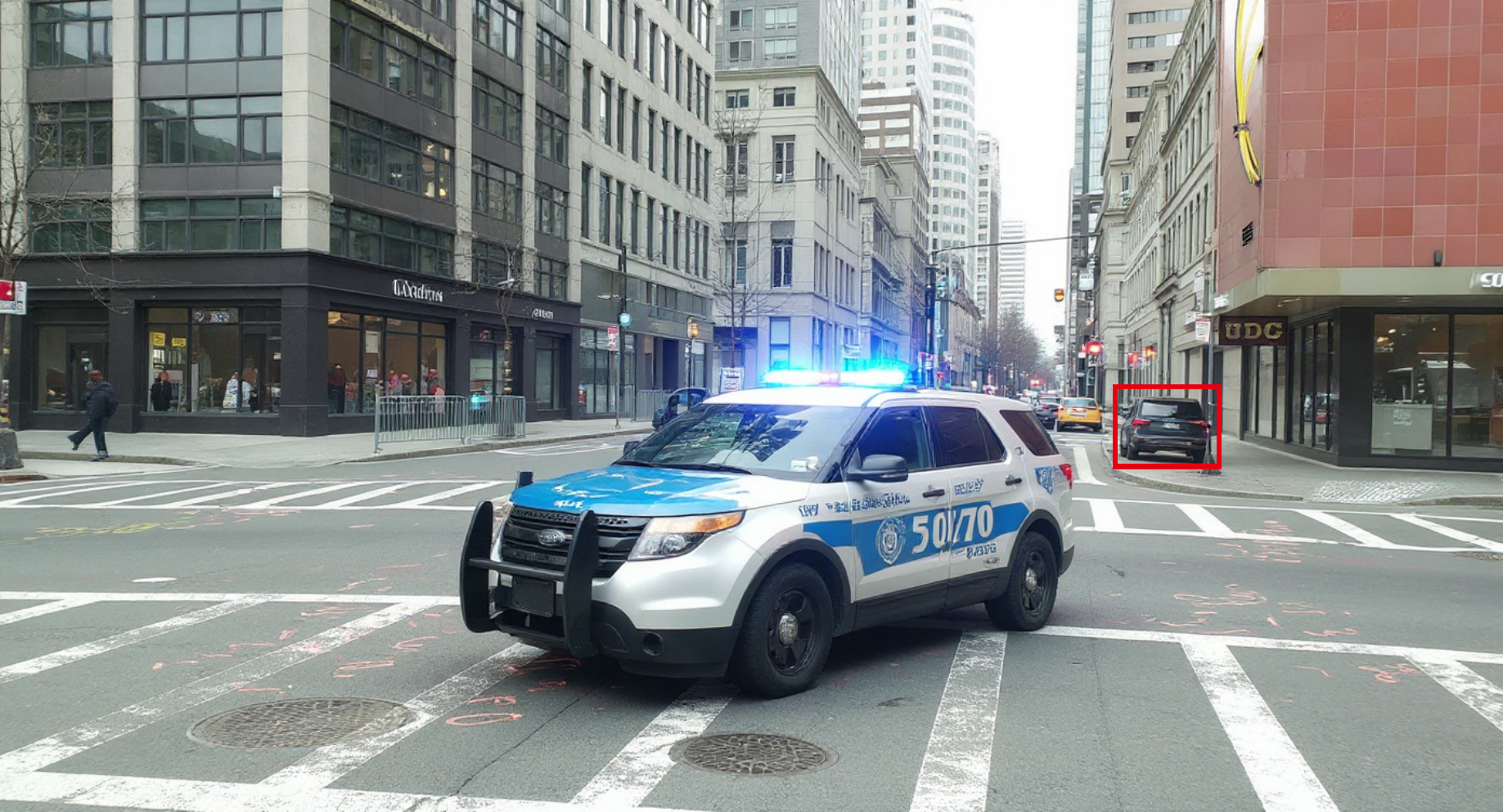} &
        \includegraphics[width=0.188\textwidth]{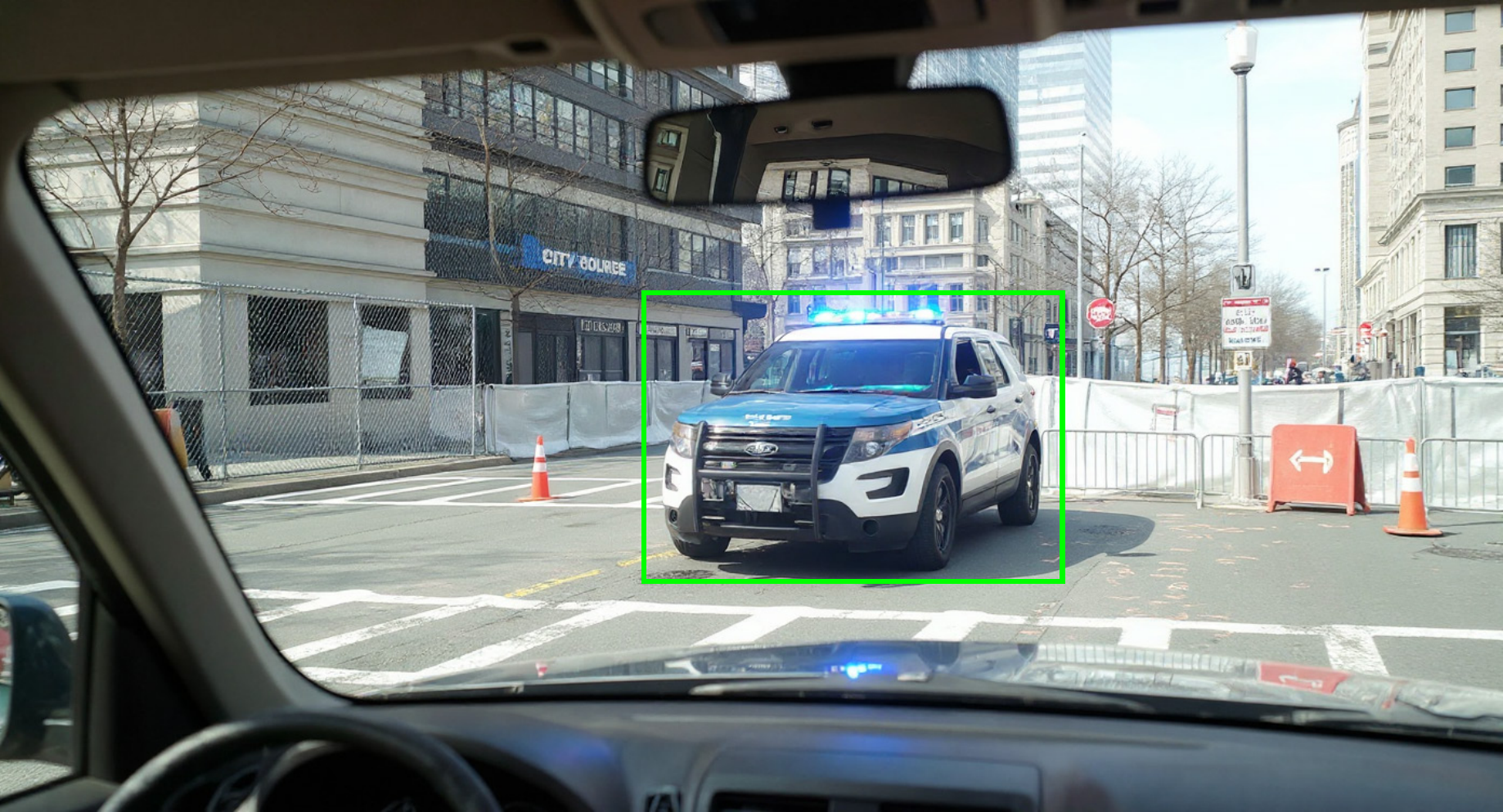} \\
        \addlinespace[2pt]
        
        \raisebox{0.7cm}{\rotatebox{90}{\scriptsize {LaRS}}} &
        \includegraphics[width=0.188\textwidth]{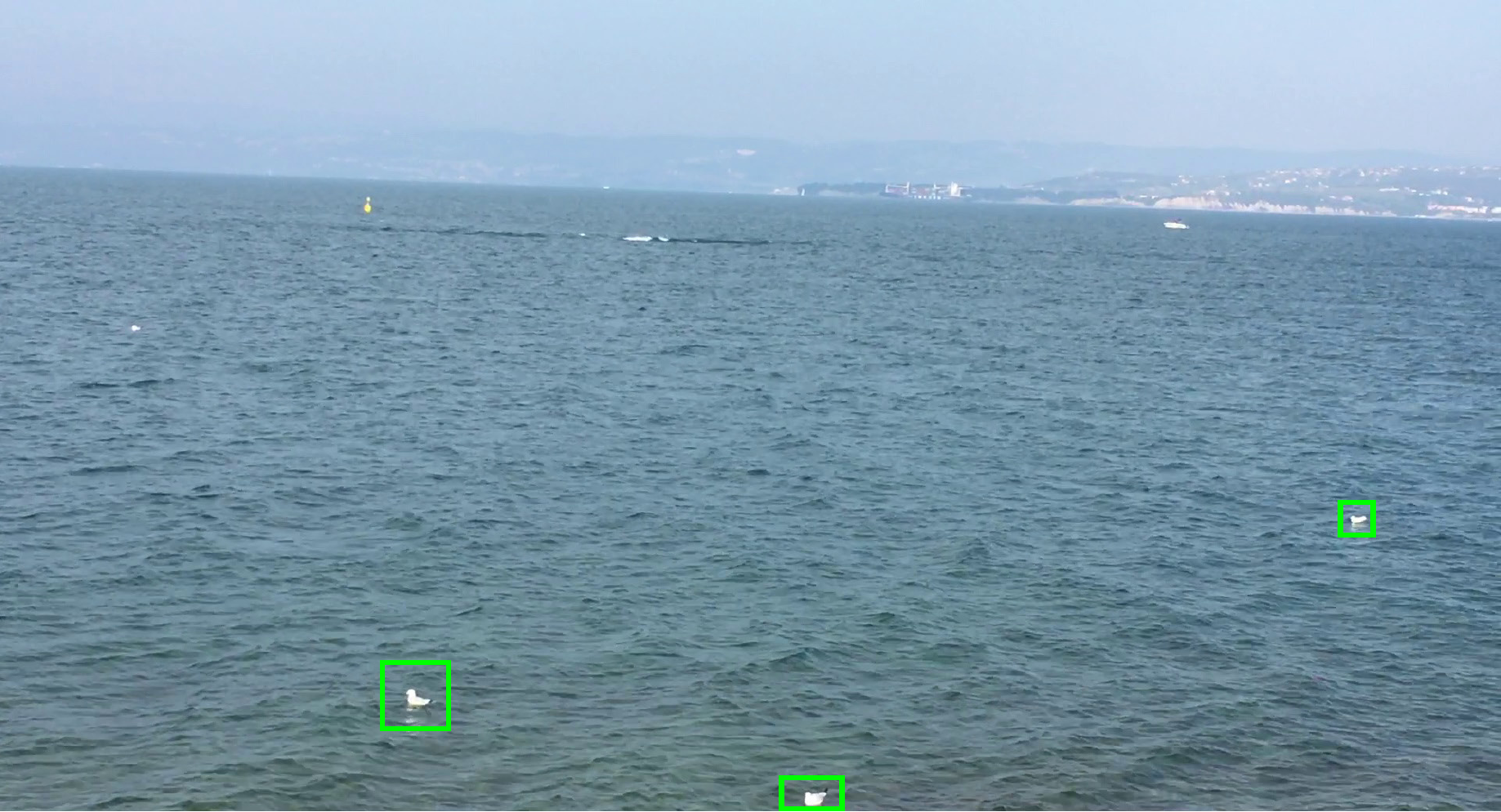} &
        \includegraphics[width=0.188\textwidth]{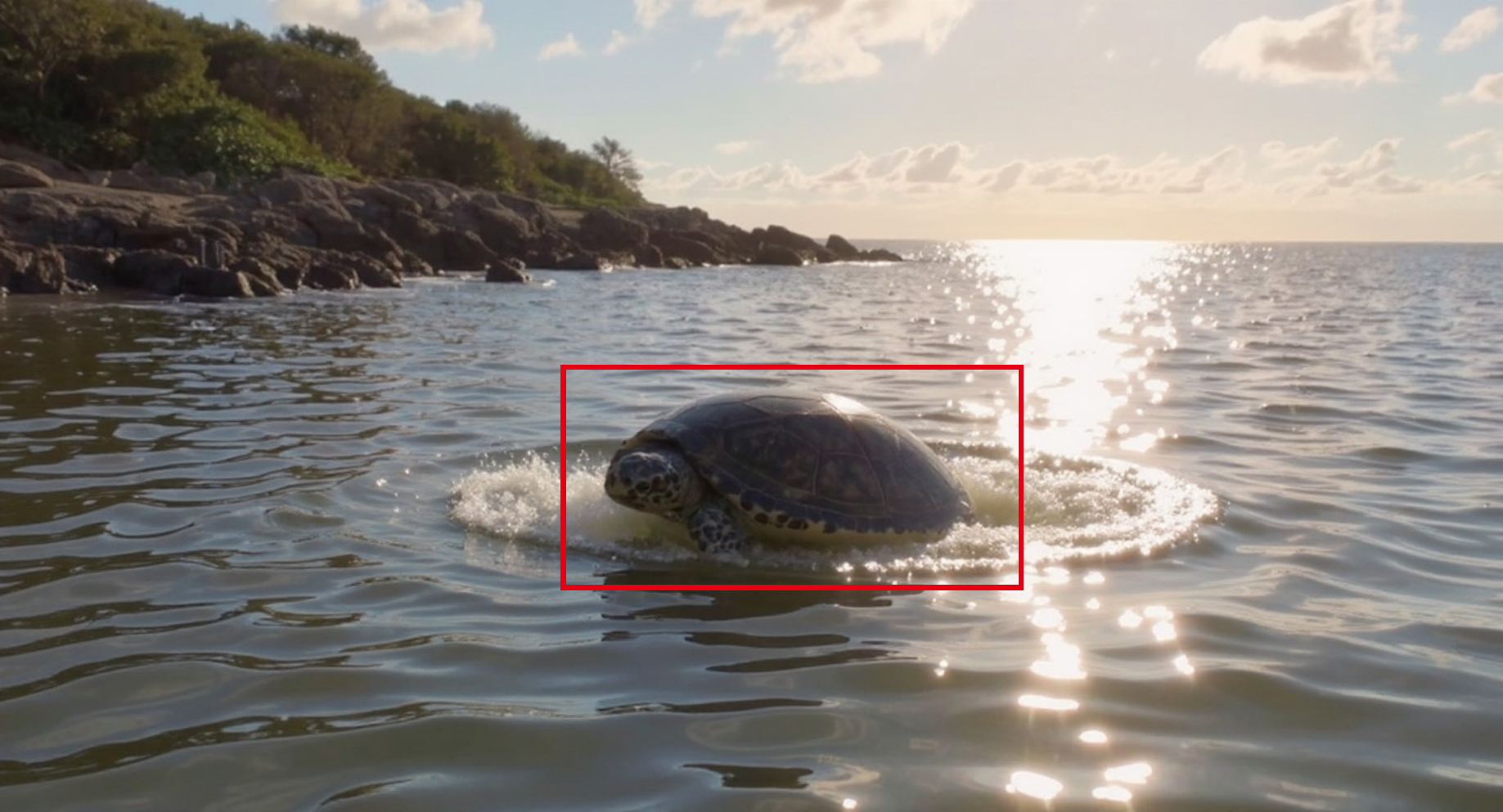} &
        \includegraphics[width=0.188\textwidth]{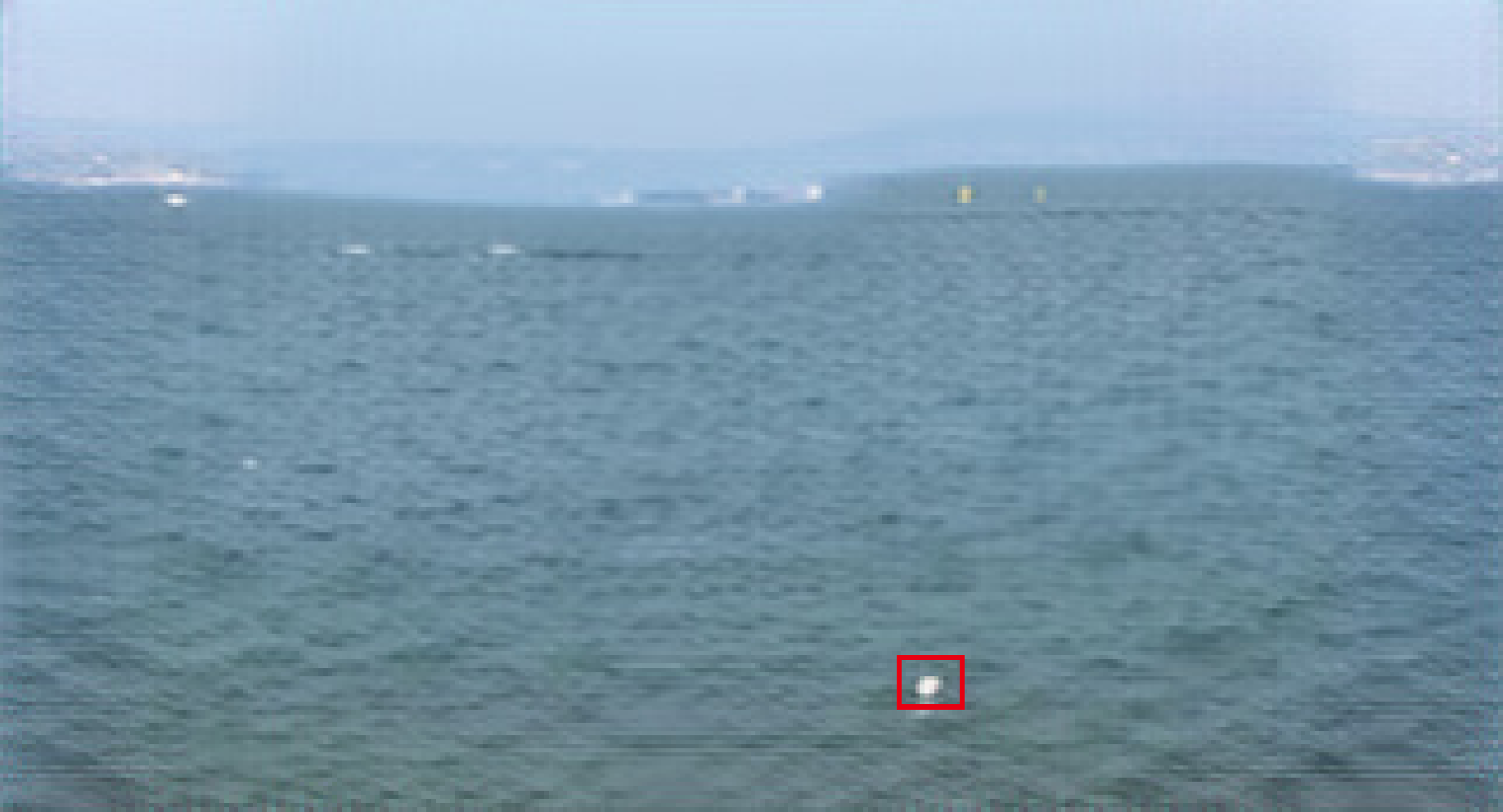} &
        \includegraphics[width=0.188\textwidth]{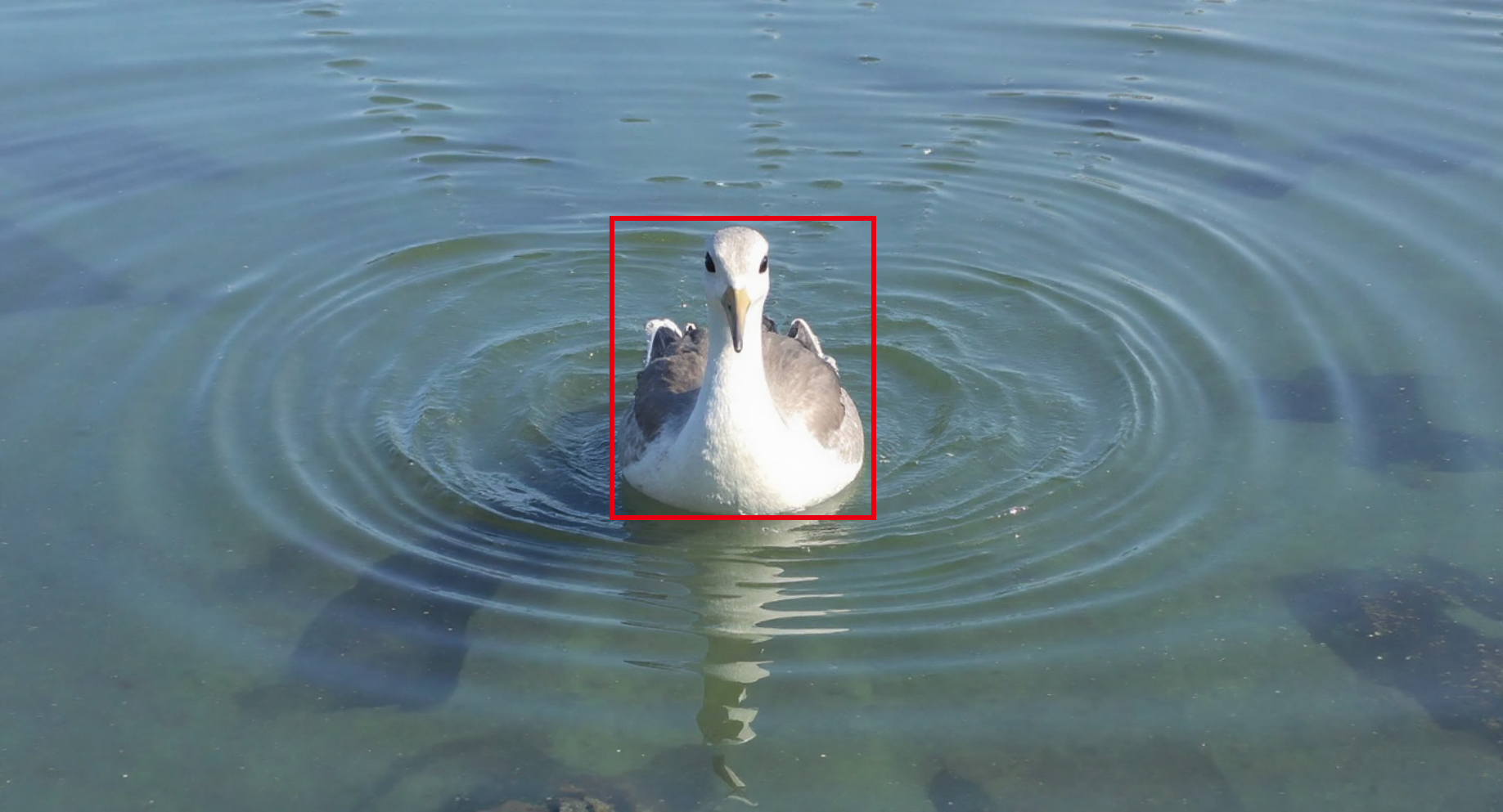} &
        \includegraphics[width=0.188\textwidth]{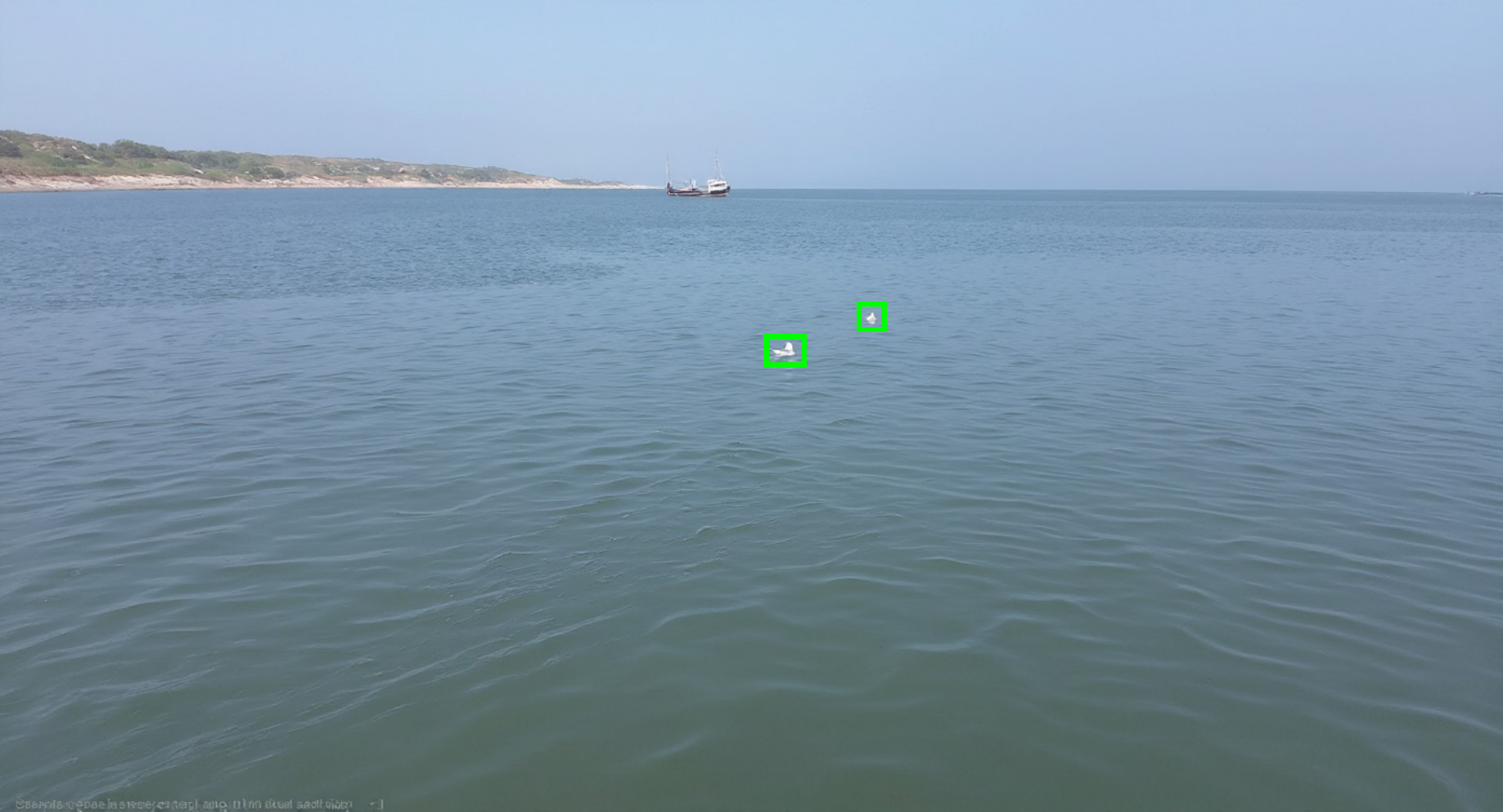} \\
    \end{tabular}
    
    \caption{Qualitative comparison of synthetic data generation across three different long-tail scenarios (rows). From left to right: the original reference image, LLM-Augmented Diffusion (Baseline 1), One-Shot GAN (Baseline 2), WMGen-v1 w/o LLM generation, and our proposed text-based world model pipeline. WMGen-v1 \rev{injects physical priors while improving spatial coherence, reducing the common hallucinations observed in the other baselines}.}
    \label{fig:qualitative_results}
\end{figure*}

Figure~\ref{fig:qualitative_results} presents qualitative comparisons between the baselines and WMGen-v1 across three representative scenarios.

\textit{1) Internal Monitoring Scenario:}
The reference image depicts a long-range surveillance view of small fishing targets on a reservoir surface. Baseline 1 exhibits strong perspective drift, generating close-range fishing scenes with large targets rather than preserving the distant monitoring viewpoint. Baseline 2 partially maintains the scene layout but produces blurred structures and distorted water–shore boundaries due to the limited representation capacity of one-shot GANs. The WMGen w/o LLM variant preserves the basic scene structure but introduces commonsense violations, such as people sitting directly in the water while fishing. In contrast, WMGen-v1 successfully expands the surrounding context while maintaining the correct surveillance perspective and realistic target scale.

\textit{2) ROADWork Scenario:}
The reference image contains a police vehicle within a road construction area. Baseline 1 frequently produces traffic-rule violations and layout inconsistencies, such as vehicles appearing on sidewalks or approaching directly from unrealistic viewpoints. Baseline 2 attempts to diversify the scene but introduces structural artifacts, such as duplicated traffic lights. The WMGen-v1 w/o LLM configuration improves the representation of the police vehicle but still generates logical inconsistencies, including vehicles appearing on sidewalks and missing the core construction-zone context. In contrast, WMGen-v1 preserves the critical construction scene while expanding the spatial layout without introducing structural or logical inconsistencies. As shown in Table~\ref{tab:prompt_comparison}, Baseline 1 mainly relies on direct prompt rewriting and provides limited explicit control over object placement, scale, and physical constraints. In contrast, WMGen-v1 organizes the prompt into structured components, including scene context, target-specific physical priors, and camera parameters. 

\textit{3) LaRS Maritime Scenario:}
The reference image shows extremely small animal targets on the ocean surface captured from an unmanned vessel viewpoint. Baseline 1 produces severe scale inconsistencies, generating large close-up animals with incorrect perspectives. Baseline 2 maintains appropriate object scales but generates overly blurred targets and exhibits limited diversity. The WMGen-v1 w/o LLM variant again suffers from scale inconsistencies similar to Baseline 1. In contrast, WMGen-v1 preserves the small-object scale and vessel viewpoint while introducing plausible variations in target positions and background conditions, resulting in more realistic maritime scenes.

\begin{table}[t]
\caption{Comparison of Prompt Structures}
\label{tab:prompt_comparison}
\centering
\begin{tabularx}{\linewidth}{@{} >{\raggedright\arraybackslash}X @{}}
\toprule
\textbf{Baseline 1 (LLM Augmented Diffusion)} \\
\midrule
Autonomous driving camera view approaching a road construction zone with cones, barriers, and lane shifts, where a marked police\_vehicle is parked on the shoulder supervising traffic control. Midday sunlight, clear sky, crisp shadows on asphalt, workers in reflective vests near excavators, realistic urban street with signage and temporary speed limit boards. \\
\midrule
\midrule
\textbf{Ours (WMGen-v1)} \\
\midrule
\textbf{[Scene Context]} \newline 
At a four-way downtown intersection, fresh asphalt patches and a fenced excavation redirect traffic into a single open lane. Orange cones form a taper toward the centerline while pedestrians wait at the crosswalk by a corner café. Overhead shop signs glow in late-afternoon sun... \vspace{1.5mm} \newline
\textbf{[Target \& Physical Prior]} \newline 
One marked police SUV with a white body and dark side panels, reflective decals, and the word ``POLICE'' on the doors, sits just upstream of the cone taper beside the barricades. Blue roof lightbar flashes brightly. \textit{It fills about 10\% of the frame in the midground, angled parallel to traffic flow, fully visible.} \vspace{1.5mm} \newline
\textbf{[Camera Parameters]} \newline 
Forward-facing in-car camera at \textit{$\sim$1.5 m height} from the ego vehicle, centered in the open lane approaching the intersection. The police SUV and construction barriers occupy the mid-to-right portion... \\
\bottomrule
\end{tabularx}
\end{table}

\subsection{Quantitative Results}
\textit{1) Overall Performance Comparison:}
All reported results are obtained using detectors trained on datasets constructed by different data generation pipelines. The detector architectures and training protocols are kept identical across all experiments, so that the observed performance differences can be attributed solely to the training data.

As shown in Table~\ref{tab:roadwork_results} and Table~\ref{tab:lars_results}, incorporating WMGen-v1-generated samples consistently improves detection performance across both YOLO and Faster R-CNN.

On the ROADWork dataset, the improvement is particularly evident for the two-stage detector. With Faster R-CNN, performance increases from 0.3602 to 0.4248 mAP@50, corresponding to an absolute gain of 0.0646 mAP@50. This result indicates that the samples generated by WMGen-v1 offer spatially informative supervision that is particularly beneficial in complex construction scenes.

On the LaRS dataset, Real + WMGen-v1 achieves the best overall results, reaching 0.5022 mAP@50 with YOLO and 0.5169 mAP@50 with Faster R-CNN. Compared with the Real-only baseline, these results correspond to absolute gains of 0.0332 and 0.0304 mAP@50, respectively. This improvement suggests that WMGen-v1 provides effective complementary training samples for rare maritime categories while maintaining plausible scene composition.

By contrast, the gains brought by Baseline 1 and Baseline 2 are limited or inconsistent across datasets and detector architectures. This observation suggests that prompt rewriting alone provides insufficient spatial grounding, while one-shot GAN-based generation has difficulty producing samples with both adequate diversity and structural consistency under extreme data scarcity.

It is also noteworthy that detectors trained solely on WMGen-v1-generated data achieve performance close to those trained only on real data. For example, on ROADWork with YOLO, WMGen-v1-only attains 0.4270 mAP@50, compared with 0.4330 for Real-only. This result indicates that WMGen-v1 can preserve task-relevant scene structure and provide effective supervision even without access to real training images.

\begin{table}[htbp]
\caption{Comparison on the ROADWork dataset.}
\label{tab:roadwork_results}
\centering
\footnotesize 
\setlength{\tabcolsep}{3 pt} 
\begin{tabular}{l c c c c}
\toprule
\multirow{2}{*}{\textbf{Method}} & \multicolumn{2}{c}{\textbf{YOLO}} & \multicolumn{2}{c}{\textbf{Faster R-CNN}} \\
\cmidrule(lr){2-3} \cmidrule(lr){4-5}
& mAP@50 & mAP@50-95 & mAP@50 & mAP@50-95 \\
\midrule
Real-only     & 0.4330 & 0.2729 & 0.3602 & 0.1926 \\
Real + Baseline 1    & 0.4328 & 0.2704 & 0.3980 & 0.2196 \\
Real + Baseline 2    & 0.4303 & 0.2721 & 0.3734 & 0.2044 \\
WMGen-v1-only     & 0.4270 & 0.2636 & 0.3779 & 0.2088 \\
\midrule
\textbf{Real + WMGen-v1}   & \textbf{0.4400} & \textbf{0.2808} & \textbf{0.4248} & \textbf{0.2420} \\
\bottomrule
\end{tabular}
\end{table}

\begin{table}[htbp]
\caption{Comparison on the LaRS dataset.}
\label{tab:lars_results}
\centering
\footnotesize 
\setlength{\tabcolsep}{3 pt} 
\begin{tabular}{l c c c c}
\toprule
\multirow{2}{*}{\textbf{Method}} & \multicolumn{2}{c}{\textbf{YOLO}} & \multicolumn{2}{c}{\textbf{Faster R-CNN}} \\
\cmidrule(lr){2-3} \cmidrule(lr){4-5}
& mAP@50 & mAP@50-95 & mAP@50 & mAP@50-95 \\
\midrule
Real-only     & 0.4690 & 0.3361 & 0.4865 & 0.3150 \\
Real + Baseline 1    & 0.4672 & 0.3356 & 0.5068 & 0.3245 \\
Real + Baseline 2    & 0.4574 & 0.3348 & 0.5065 & 0.3315 \\
WMGen-v1-only     & 0.4487 & 0.3155 & 0.4607 & 0.2830 \\
\midrule
\textbf{Real + WMGen-v1}   & \textbf{0.5022} & \textbf{0.3539} & \textbf{0.5169} & \textbf{0.3335} \\
\bottomrule
\end{tabular}
\end{table}

\textit{2) Per-class Analysis on the ROADWork Dataset:}
Table~\ref{tab:roadwork_per_class} reports the per-class mAP@50 results on the ROADWork dataset using YOLO. Our data generation pipeline is designed to augment several sparse and safety-relevant categories, including Police Officer, Police Vehicle, Other Roadwork Objects, and Bike Lane.

WMGen-v1 yields clear improvements for several of these underrepresented categories. In particular, the mAP@50 for Other Roadwork Objects increases from 0.0009 to 0.1310, corresponding to an absolute gain of 0.1301, while Police Officer improves from 0.1060 to 0.1990, corresponding to an absolute gain of 0.0930. Smaller but consistent gains are also observed for Bike Lane, which increases from 0.4680 to 0.4830. These results suggest that WMGen-v1 is particularly effective for categories with limited training samples and relatively flexible spatial configurations.

A notable exception is Police Vehicle, whose mAP@50 decreases from 0.2760 to 0.2580. One possible explanation is that this category depends more heavily on accurate shape regularity, fine-grained local texture, and category-specific visual cues. Although WMGen-v1 can constrain coarse scene placement through language-guided priors, the generated samples may still contain geometric inconsistencies or texture deviations under challenging viewpoints. In addition, in a multi-object detection setting, police vehicles may co-occur with ordinary vehicles in the same scene, making the category boundary more ambiguous when distinctive visual cues are not rendered clearly. Such ambiguity may increase classification confusion and reduce the benefit of augmentation for this class.

Overall, the per-class results indicate that WMGen-v1 is most beneficial for sparse categories whose recognition relies primarily on contextual placement and coarse semantic appearance, while its benefit is more limited for highly structured rigid objects requiring fine-grained geometric fidelity.

\begin{table}[t]
\caption{Per-class mAP@50 Results on the ROADWork Dataset}
\label{tab:roadwork_per_class}
\centering
\scriptsize 
\setlength{\tabcolsep}{1.5 pt} 
\begin{tabular}{l c c c c}
\toprule
\textbf{Class} & \textbf{Real-only} & \textbf{Baseline 1} & \textbf{Baseline 2} & \textbf{Real + WMGen-v1} \\
\midrule
\textbf{Police Officer} & 0.1060 & 0.1110 & 0.0737 & \textbf{0.1990} \\
\textbf{Police Vehicle} & \textbf{0.2760} & 0.2560 & 0.2720 & 0.2580 \\
Cone & 0.6930 & 0.6890 & 0.6930 & 0.6910 \\
Fence & 0.3390 & 0.3390 & 0.3530 & 0.3620 \\
Drum & 0.7210 & 0.7220 & 0.7180 & 0.7090 \\
Barricade & 0.5330 & 0.5280 & 0.5100 & 0.5130 \\
Barrier & 0.4070 & 0.4090 & 0.4280 & 0.4120 \\
Work Vehicle & 0.4790 & 0.4950 & 0.4780 & 0.4890 \\
Vertical Panel & 0.6050 & 0.6170 & 0.6210 & 0.6160 \\
Tubular Marker & 0.5980 & 0.6070 & 0.6050 & 0.6010 \\
Arrow Board & 0.7090 & 0.6920 & 0.7100 & 0.7060 \\
\textbf{Bike Lane} & 0.4680 & 0.4430 & 0.3540 & \textbf{0.4830} \\
Work Equipment & 0.0474 & 0.0403 & 0.0662 & 0.0689 \\
Worker & 0.4220 & 0.4380 & 0.4180 & 0.4110 \\
\textbf{Other Roadwork Objects} & 0.0009 & 0.0055 & 0.0027 & \textbf{0.1310} \\
Temp Traffic Msg Board & 0.4440 & 0.4530 & 0.4940 & 0.3810 \\
Temp Traffic Sign & 0.5180 & 0.5140 & 0.5280 & 0.5160 \\
\midrule
\textbf{mAP@50 (All)} & 0.4330 & 0.4330 & 0.4310 & \textbf{0.4440} \\
\bottomrule
\end{tabular}
\end{table}

\textit{3) Per-class Analysis on the LARS Dataset:}
Table~\ref{tab:lars_per_class} reports the per-class mAP@50 results on the LARS dataset using YOLO. On this maritime benchmark, WMGen-v1 is mainly used to augment several sparse categories, including Float and Animal.

WMGen-v1 improves multiple underrepresented categories. In particular, Float increases from 0.0000 to 0.3840, indicating that the generated samples provide useful supervision for a category that is nearly absent in the original training distribution. Animal also improves from 0.0837 to 0.1980, corresponding to an absolute gain of 0.1143. These results suggest that WMGen-v1 is effective for sparse maritime obstacles whose recognition depends primarily on coarse semantic appearance and contextual placement.

Meanwhile, the performance of several non-augmented frequent categories, such as Water and Sky, remains broadly stable, while structurally constrained object classes, such as Boat/ship, show limited benefit. \revz{Among these, swimmer shows a noticeable AP drop after adding Animal/Float samples. Supplementary diagnostics suggest that this behavior is associated with confusion among visually similar small water-surface objects rather than sample scarcity alone.} Overall, the per-class results indicate that WMGen-v1 is particularly beneficial for sparse categories with limited training coverage, whereas its effect is more constrained for categories requiring precise geometric structure or fine-grained water--object interaction modeling.

\begin{table}[htbp]
\caption{Per-class mAP@50 Results on the LARS Dataset}
\label{tab:lars_per_class}
\centering
\footnotesize 
\setlength{\tabcolsep}{1.5 pt} 
\begin{tabular}{l c c c c}
\toprule
\textbf{Class} & \textbf{Real-only} & \textbf{Baseline 1} & \textbf{Baseline 2} & \textbf{Real + WMGen-v1} \\
\midrule
Static Obstacle & 0.6570 & 0.6730 & 0.7120 & 0.6810 \\
Water & 0.9940 & 0.9940 & 0.9950 & 0.9930 \\
Sky & 0.9840 & 0.9670 & 0.9810 & 0.9510 \\
Boat/ship & 0.5750 & 0.5620 & 0.5500 & 0.5440 \\
Row boats& 0.4460 & 0.3980 & 0.4630 & 0.4410 \\
Paddle board & 0.5500 & 0.6390 & 0.5320 & 0.5470 \\
Buoy & 0.3070 & 0.3270 & 0.3250 & 0.3600 \\
Swimmer & 0.3590 & 0.4080 & 0.2830 & 0.2070 \\
\textbf{Animal} & 0.0837 & 0.0122 & 0.0375 & \textbf{0.1980} \\
\textbf{Float} & 0.0000 & 0.0000 & 0.0000 & \textbf{0.3840} \\
Other & 0.2070 & 0.1570 & 0.1490 & 0.2180 \\
\midrule
\textbf{mAP@50 (All)} & 0.4690 & 0.4670 & 0.4570 & \textbf{0.5020} \\
\bottomrule
\end{tabular}
\end{table}

\textit {4) Long-Tail Category Analysis on both the ROADWork and LaRS Datasets:}
We define rare long-tail classes as those with fewer than 100 training images in the original dataset and report their averaged performance separately.

Under this definition, the ROADWork dataset contains four long-tail categories (\textit{Police Officer}, \textit{Police Vehicle}, \textit{Bike Lane}, and \textit{Other Roadwork Objects}). As shown in Table~\ref{tab:longtail_per_class}, The average mAP of these classes increases from 0.213 under the Real Only baseline to 0.268 when incorporating our generated data, corresponding to an absolute gain of 0.055. Similarly, the LaRS dataset contains two long-tail categories (\textit{Animal} and \textit{Float}). The average mAP increases from 0.0419 to 0.2910 with our method, corresponding to an absolute gain of 0.2491. When aggregating all six long-tail categories across both datasets, the overall mAP increases from 0.1558 to 0.2755, corresponding to an absolute gain of 0.1197. 

Most individual long-tail categories also exhibit consistent performance improvements compared with the baseline. These results demonstrate that our world-model-driven data generation framework is particularly effective at enriching rare categories, alleviating data scarcity, and improving detection performance on long-tail classes.

\begin{table}[htbp]
\caption{Per-class mAP@50 Results on Long-Tail Categories.}
\label{tab:longtail_per_class}
\centering
\scriptsize 
\setlength{\tabcolsep}{6pt} 
\begin{tabular}{l c c c c}
\toprule
\textbf{Long-Tail Class} &
\makecell[c]{\textbf{Real} \\ \textbf{Images}} & 
\makecell[c]{\textbf{Generated} \\ \textbf{Images}} & 
\makecell[c]{\textbf{Real} \\ \textbf{Only}} & 
\makecell[c]{\textbf{Real +} \\ \textbf{WMGen-v1}} \\
\midrule
Police Officer & 69 & 100 & 0.1060 & \textbf{0.1990} \\
Police Vehicle & 87 & 100 & \textbf{0.2760} & 0.2580 \\
Bike Lane & 61 & 40 & 0.4680 & \textbf{0.4830} \\
Other Roadwork Objects & 48 & 40 & 0.0009 & \textbf{0.1310} \\
Animal & 72 & 40 & 0.0837 & \textbf{0.1980} \\
Float & 20 & 20 & 0.0000 & \textbf{0.3840} \\
\midrule
\textbf{mAP@50 (Avg)} & -- & -- & 0.1558 & \textbf{0.2755} \\
\bottomrule
\end{tabular}
\end{table}

\section{CONCLUSION}
In this paper, we proposed WMGen-v1, an agentic text-based world model for long-tail visual perception. WMGen-v1 generates physically grounded training samples from a single user input and reference image. \rev{By explicitly injecting visual and physical priors, WMGen-v1 reduces common semantic drift, layout violations, and scale hallucinations observed in the evaluated baselines.}

Experiments on industrial, ROADWork, and LaRS datasets demonstrate that WMGen-v1 consistently enhances downstream detection for both YOLO and Faster R-CNN. \revz{Combining real and synthetic data yields the best overall performance, effectively mitigating perception blind spots in rare, safety-critical scenarios and expanding the safe operational envelope of autonomous systems.}

Despite these promising results, limitations remain. Converting visual observations into discrete text descriptions can cause spatial information loss, particularly affecting objects that require precise geometric structures or complex physical interactions. \revz{The per-class results suggest that class frequency alone is insufficient for selecting augmentation targets. Moderate-frequency classes (e.g., Work Equipment) can remain challenging, while visually ambiguous non-target classes (e.g., Swimmer) may be affected by augmentation of related categories.} Furthermore, the current open-loop pipeline lacks a feedback mechanism to verify the physical and semantic correctness of the generated scenes.


\balance
\bibliographystyle{IEEEtran}
\bibliography{refs}

@article{chen2024geo,
  author = {Chen, Kai and Xie, Enze and Chen, Zhe and Wang, Yibo and Hong, Lanqing and Li, Zhenguo and Yeung, Dit-Yan},
  id = {chen2024geo},
  journal = {Published as a conference paper at ICLR 2024},
  title = {Geodiffusion: Text-prompted geometric control for object detection data generation},
  year = {2024}
}

@article{chen2026spa,
  author = {Chen, Wei and Long, Yancheng and Liu, Mingqiao and Ding, Haojie and Yang, Yankai and Wei, Hongyang and Zhang, Yi-Fan and Wen, Bin and Yang, Fan and Gao, Tingting and Li, Han and Chen, Long},
  id = {chen2026spa},
  title = {Spatial Chain-of-Thought: Bridging Understanding and Generation Models for Spatial Reasoning Generation},
  journal={arXiv preprint arXiv:2602.11980},
  url = {https://weichens.github.io/spatial_chain_of_thought/},
  year = {2026}
}

@article{cong2024dec,
  author = {Cong, Cong and Xuan, Shiyu and Liu, Sidong and Zhang, Shiliang and Pagnucco, Maurice and Song, Yang},
  doi = {10.1609/aaai.v38i2.27902},
  id = {cong2024dec},
  journal = {Proceedings of the AAAI Conference on Artificial Intelligence},
  month = {mar},
  number = {2},
  pages = {1380--1388},
  publisher = {Association for the Advancement of Artificial Intelligence (AAAI)},
  title = {Decoupled Optimisation for Long-Tailed Visual Recognition},
  url = {http://dx.doi.org/10.1609/aaai.v38i2.27902},
  volume = {38},
  year = {2024}
}

@article{du2024pro,
  author = {Du, Chaoqun and Wang, Yulin and Song, Shiji and Huang, Gao},
  doi = {10.1109/tpami.2024.3369102},
  id = {du2024pro},
  issn = {1939-3539},
  journal = {IEEE Transactions on Pattern Analysis and Machine Intelligence},
  month = {sep},
  number = {9},
  pages = {5890--5904},
  publisher = {Institute of Electrical and Electronics Engineers (IEEE)},
  title = {Probabilistic Contrastive Learning for Long-Tailed Visual Recognition},
  url = {http://dx.doi.org/10.1109/TPAMI.2024.3369102},
  volume = {46},
  year = {2024}
}

@inproceedings{fang2024dat,
  author = {Fang, Haoyang and Han, Boran and Zhang, Shuai and Zhou, Su and Hu, Cuixiong and Ye, Wen-Ming},
  id = {fang2024dat},
  title = {Data Augmentation for Object Detection via Controllable Diffusion Models},
  year = {2024},
  booktitle={Proceedings of the IEEE/CVF winter conference on applications of computer vision}
}

@inproceedings{fu2024bli,
  author = {Fu, Xingyu and Hu, Yushi and Li, Bangzheng and Feng, Yu and Wang, Haoyu and Lin, Xudong and Roth, Dan and Smith, Noah A. and Ma, Wei-Chiu and Krishna, Ranjay},
  doi = {10.48550/ARXIV.2404.12390},
  id = {fu2024bli},
  publisher = {arXiv},
  title = {BLINK: Multimodal Large Language Models Can See but Not Perceive},
  url = {https://arxiv.org/abs/2404.12390},
  year = {2024},
  booktitle={European Conference on Computer Vision}
}

@inproceedings{ghosh2023roa,
  author = {Ghosh, Anurag and Zheng, Shen and Tamburo, Robert and Vuong, Khiem and Alvarez-Padilla, Juan and Zhu, Hailiang and Cardei, Michael and Dunn, Nicholas and Mertz, Christoph and Narasimhan, Srinivasa G.},
  id = {ghosh2023roa},
  title = {ROADWork: A Dataset and Benchmark for Learning to Recognize, Observe, Analyze and Drive Through Work Zones},
  booktitle={Proceedings of the IEEE/CVF International Conference on Computer Vision},
  url = {https://www.cs.cmu.edu/~roadwork/},
  year = {2023}
}

@article{guo2025srm,
  author = {Guo, Yulong and Zhang, Zilun and Shang, Yongheng and Zhao, Tiancheng and Deng, Shuiguang and Yang, Yingchun and Yin, Jianwei},
  doi = {10.1109/tgrs.2025.3565600},
  id = {guo2025srm},
  issn = {1558-0644},
  journal = {IEEE Transactions on Geoscience and Remote Sensing},
  pages = {1--16},
  publisher = {Institute of Electrical and Electronics Engineers (IEEE)},
  title = {SRMF: A Data Augmentation and Multimodal Fusion Approach for Long-Tail UHR Satellite Image Segmentation},
  url = {http://dx.doi.org/10.1109/TGRS.2025.3565600},
  volume = {63},
  year = {2025}
}

@inbook{hyuncho2022lon,
  author = {Hyun Cho, Jang and Krähenbühl, Philipp},
  booktitle = {Computer Vision - ECCV 2022},
  doi = {10.1007/978-3-031-20074-8_40},
  id = {hyuncho2022lon},
  isbn = {9783031200748},
  issn = {1611-3349},
  pages = {698--714},
  publisher = {Springer Nature Switzerland},
  title = {Long-tail Detection with Effective Class-Margins},
  url = {http://dx.doi.org/10.1007/978-3-031-20074-8_40},
  year = {2022}
}

@article{li2026min,
  author = {Li, Yanwei and Zhang, Yuechen and Wang, Chengyao and Zhong, Zhisheng and Chen, Yixin and Chu, Ruihang and Liu, Shaoteng and Jia, Jiaya},
  doi = {10.1109/tpami.2025.3637265},
  id = {li2026min},
  issn = {1939-3539},
  journal = {IEEE Transactions on Pattern Analysis and Machine Intelligence},
  month = {mar},
  number = {3},
  pages = {3530--3543},
  publisher = {Institute of Electrical and Electronics Engineers (IEEE)},
  title = {Mini-Gemini: Mining the Potential of Multi-Modality Vision Language Models},
  url = {http://dx.doi.org/10.1109/TPAMI.2025.3637265},
  volume = {48},
  year = {2026}
}

@article{pernias2023wue,
  author = {Pernias, Pablo and Rampas, Dominic and Richter, Mats L. and Pal, Christopher J. and Aubreville, Marc},
  journal={arXiv preprint arXiv:2306.00637},
  copyright = {Creative Commons Attribution 4.0 International},
  doi = {10.48550/ARXIV.2306.00637},
  id = {pernias2023wue},
  publisher = {arXiv},
  title = {Wuerstchen: An Efficient Architecture for Large-Scale Text-to-Image Diffusion Models},
  url = {https://arxiv.org/abs/2306.00637},
  year = {2023}
}

@misc{podell2023sdx,
  author = {Podell, Dustin and English, Zion and Lacey, Kyle and Blattmann, Andreas and Dockhorn, Tim and Müller, Jonas and Penna, Joe and Rombach, Robin},
  copyright = {Creative Commons Attribution 4.0 International},
  doi = {10.48550/ARXIV.2307.01952},
  id = {podell2023sdx},
  publisher = {arXiv},
  title = {SDXL: Improving Latent Diffusion Models for High-Resolution Image Synthesis},
  url = {https://arxiv.org/abs/2307.01952},
  year = {2023}
}

@inproceedings{sushko2021sho,
  author = {Sushko, Vadim and Gall, Jürgen and Khoreva, Anna},
  booktitle = {Proceedings of the IEEE/CVF Conference on Computer Vision and Pattern Recognition (CVPR)},
  id = {sushko2021sho},
  title = {One-Shot GAN: Learning to Generate Samples from Single Images and Videos},
  year = {2021}
}

@inproceedings{tan2020equ,
  author = {Tan, Jingru and Wang, Changbao and Li, Buyu and Li, Quanquan and Ouyang, Wanli and Yin, Changqing and Yan, Junjie},
  booktitle = {2020 IEEE/CVF Conference on Computer Vision and Pattern Recognition (CVPR)},
  doi = {10.1109/cvpr42600.2020.01168},
  id = {tan2020equ},
  month = {jun},
  pages = {11659--11668},
  publisher = {IEEE},
  title = {Equalization Loss for Long-Tailed Object Recognition},
  url = {http://dx.doi.org/10.1109/cvpr42600.2020.01168},
  year = {2020}
}

@article{ust2023lar,
  author = {Žust, Lojze and Pers, Janez and Kristan, Matej},
  id = {ust2023lar},
  journal = {arXiv preprint arXiv:2308.09618},
  title = {LaRS: A Diverse Panoptic Maritime Obstacle Detection Dataset and Benchmark},
  year = {2023}
}

@misc{xia2023llm,
  author = {Xia, Bin and Wang, Shiyin and Tao, Yingfan and Wang, Yitong and Jia, Jiaya},
  copyright = {Creative Commons Attribution 4.0 International},
  doi = {10.48550/ARXIV.2311.16500},
  id = {xia2023llm},
  publisher = {arXiv},
  title = {LLMGA: Multimodal Large Language Model based Generation Assistant},
  url = {https://arxiv.org/abs/2311.16500},
  year = {2023}
}

@misc{xu2024lla,
  author = {Xu, Guowei and Jin, Peng and Wu, Ziang and Li, Hao and Song, Yibing and Sun, Lichao and Yuan, Li},
  copyright = {arXiv.org perpetual, non-exclusive license},
  doi = {10.48550/ARXIV.2411.10440},
  id = {xu2024lla},
  publisher = {arXiv},
  title = {LLaVA-CoT: Let Vision Language Models Reason Step-by-Step},
  url = {https://arxiv.org/abs/2411.10440},
  year = {2024}
}

@article{yang2023lon,
  author = {Yang, Jiaxin and Yu, Miaomiao and Li, Shuohao and Zhang, Jun and Hu, Shengze},
  id = {yang2023lon},
  journal = {Remote Sens.},
  number = {4539},
  pages = {https://doi.org/10.3390/rs15184539},
  title = {Long-Tailed Object Detection for Multimodal Remote Sensing Images},
  volume = {15},
  year = {2023}
}

@article{labs2025flux,
  title={FLUX. 1 Kontext: Flow Matching for In-Context Image Generation and Editing in Latent Space},
  author={Labs, Black Forest and Batifol, Stephen and Blattmann, Andreas and Boesel, Frederic and Consul, Saksham and Diagne, Cyril and Dockhorn, Tim and English, Jack and English, Zion and Esser, Patrick and others},
  journal={arXiv preprint arXiv:2506.15742},
  year={2025}
}

@article{hassani2026yolo,
  title={From YOLO V1 to YOLO V11: comparative analysis of YOLO algorithm},
  author={Hassani, Imane Beqqali and Benhida, Soufia and Lamii, Nabil and Oqaidi, Khalid and Ouiddad, Ahmed and Ghiadi, Soukaina},
  journal={International Journal of Electrical and Computer Engineering (IJECE)},
  volume={16},
  number={1},
  pages={450--462},
  year={2026}
}

@article{ren2015faster,
  title={Faster r-cnn: Towards real-time object detection with region proposal networks},
  author={Ren, Shaoqing and He, Kaiming and Girshick, Ross and Sun, Jian},
  journal={Advances in neural information processing systems},
  volume={28},
  year={2015}
}

@article{chen2015microsoft,
  title={Microsoft coco captions: Data collection and evaluation server},
  author={Chen, Xinlei and Fang, Hao and Lin, Tsung-Yi and Vedantam, Ramakrishna and Gupta, Saurabh and Doll{\'a}r, Piotr and Zitnick, C Lawrence},
  journal={arXiv preprint arXiv:1504.00325},
  year={2015}
}

@inproceedings{liu2024grounding,
  title={Grounding dino: Marrying dino with grounded pre-training for open-set object detection},
  author={Liu, Shilong and Zeng, Zhaoyang and Ren, Tianhe and Li, Feng and Zhang, Hao and Yang, Jie and Jiang, Qing and Li, Chunyuan and Yang, Jianwei and Su, Hang and others},
  booktitle={European conference on computer vision},
  pages={38--55},
  year={2024},
  organization={Springer}
}

\end{document}